\documentclass[runningheads]{llncs}

\usepackage{eccv}

\usepackage{graphicx}
\usepackage{booktabs}
\usepackage{mathtools}
\usepackage{tikz}
\usepackage{multirow}
\usepackage{array}
\usepackage{placeins} % preamble
\usepackage{makecell}
\usetikzlibrary{arrows.meta,positioning,fit,calc,backgrounds}
\usepackage{enumerate}

\usepackage[breaklinks,colorlinks,citecolor=eccvblue,linkcolor=eccvblue,urlcolor=eccvblue]{hyperref}
\usepackage{fontawesome5}
\usepackage{orcidlink}
\usepackage{marvosym}
\newcommand{\correspondingauthoricon}{\textsuperscript{(\Letter)}}

\newcommand{\R}{\mathbb{R}}
\newcommand{\Z}{\mathbb{Z}}
\newcommand{\x}{\mathbf{x}}
\newcommand{\vcoord}{\mathbf{v}}
\newcommand{\kclick}{\mathbf{k}}
\newcommand{\feat}{\mathbf{f}}
\newcommand{\Hh}{\mathbf{H}}
\newcommand{\Pp}{\mathbf{P}}
\newcommand{\Tt}{\mathbf{T}}
\newcommand{\slog}{\mathbf{s}}
\newcommand{\sigmoid}{\sigma}

\begin{document}

% ------------------------------------------------------------
% Title + author:
% - In review mode, eccv.sty overrides author/institute automatically
% - In final mode, fill in real author/institute below
% ------------------------------------------------------------
\title{NegROI: Click-Centric Uncertainty-Guided Refinement with Scene-Conditioned Negative Prompts for Robust Interactive 3D Segmentation}
\titlerunning{NegROI for Robust Interactive 3D Segmentation}

% Camera-ready author information:
% Feng Wu is the corresponding author. ORCID identifiers are shown as inline icons.
\author{Shuheng Zhang\orcidlink{0009-0008-5368-692X} \and
Feng Wu\correspondingauthoricon\thanks{Feng Wu is the corresponding author.}\orcidlink{0000-0003-3989-0509}}
\authorrunning{S. Zhang and F. Wu}
\institute{School of Computer Science and Technology, \\ University of Science and Technology of China, Hefei, Anhui, China.\\
\email{zsh123456@mail.ustc.edu.cn}, \email{wufeng02@ustc.edu.cn}
}

\maketitle
% ============================================================
\begin{abstract}
Interactive 3D segmentation aims to extract object masks in point clouds with minimal user clicks.
Despite recent progress, most existing approaches still struggle with (i) coarse voxel resolution that blurs fine boundaries under limited clicks and (ii) hard false positives caused by confusing background structures.
These issues are exacerbated by density and scale shifts across datasets (e.g., dense RGB-D reconstructions vs. sparse LiDAR scans), where fixed refinement heuristics and purely click-driven decoding generalize poorly.
To address them, we propose NegROI --- a novel transformer-based interactive framework that couples click-centric multi-resolution refinement with scene-conditioned negative prompts.
Given a coarse voxel prediction, it refines only a local Region Of Interest (ROI) around the current click on a finer grid and fuses refined logits back to the coarse mask.
To improve robustness and efficiency, we introduce uncertainty-driven selective refinement that prioritizes ambiguous regions.
Meanwhile, we model hard background patterns via a set of scene-conditioned negative prompts obtained by cross-attention over scene tokens. We further stabilize these prompts with a diversity regularizer.
Finally, we propose boundary-aware hard negative mining to supervise negative-prompt attention toward boundary-proximal, high-confidence false positives.
Our experiments on common benchmark datasets (i.e., ScanNet, S3DIS, and KITTI) demonstrate improved click efficiency and reduced false positives, with stronger cross-dataset robustness than the state-of-the-art baselines.
\end{abstract}

% ============================================================
\section{Introduction}
Interactive segmentation reduces annotation cost by iteratively refining a mask with user feedback.
In many 3D applications, interactive segmentation is particularly valuable because dense point-wise labels are often expensive to acquire.
However, point clouds exhibit large variability in density, noise, and scale. Therefore, interactive pipelines must respond reliably to sparse user clicks.

Modern interactive 3D segmentation methods \cite{kontogianni2022interactive,yue2023agile3d,simonelli2025easy3d,WHZWtmm26} commonly operate on voxel grids for efficiency and use transformer decoders to integrate click tokens with scene features.
While effective, we observe two persistent failure modes:
(1) \textit{Boundary under-segmentation:} A single-resolution voxel representation tends to blur object boundaries, especially when the click budget is small.
(2) \textit{Hard false positives (FPs):} Even with clicks, confusing background structures can trigger high-confidence FPs.
Both issues are amplified under distribution shifts (e.g., ScanNet-trained models tested on S3DIS or KITTI), where background appearance and sampling patterns differ substantially.

Against this background, we propose  a novel method named \textit{NegROI}, which targets both issues with a unified design.
Firstly, we introduce \emph{click-centric multi-resolution refinement}: after a coarse prediction, we refine only a local ROI around the \emph{current} click on a finer grid and fuse the refined prediction back to the coarse mask.
Secondly, we introduce \emph{scene-conditioned negative prompts} that act as background prototypes learned by cross-attending to scene tokens, providing a structured way to suppress hard FPs.
To improve robustness and efficiency, we add uncertainty-driven selective refinement, a diversity regularizer to prevent prompt collapse, and boundary-aware hard negative mining to directly supervise negative-prompt attention on boundary-adjacent confusing background.

%\paragraph{Contributions.}
Our main contributions are summarized as follows:
\begin{enumerate}[(1)]
  \item We propose the \textit{click-centric multi-resolution refinement} module that performs efficient fine-grid refinement inside a local ROI and fuses fine predictions back to coarse logits.
  \item We propose the \textit{uncertainty-driven selective refinement} to focus refinement computation on ambiguous regions, improving efficiency and robustness.
  \item We propose the \textit{scene-conditioned negative prompts} for explicit background modeling, augmented with a \textit{prompt diversity regularizer}.
  \item We propose the \textit{boundary-aware hard negative mining} to supervise negative-prompt attention toward boundary-proximal, high-confidence false positives.
\end{enumerate}
All together, we advance the state-of-the-art for robust interactive 3D segmentation. The experimental results on common benchmark datasets (i.e., ScanNet, S3DIS, and KITTI) demonstrate  im-
proved click efficiency, reduced false positives, and cross-dataset robustness of our approach. 

% ============================================================
\section{Related Work}
\label{sec:related}

In this section, we briefly review the related work on interactive 3D segmentation.

\paragraph{Interactive image segmentation.}
Early interactive segmentation methods leverage sparse user cues such as extreme points \cite{maninis2018deep} and iterative refinement strategies \cite{sofiiuk2020f,sofiiuk2022reviving}.
Recent work improves click efficiency by better modeling click locality and hard errors (e.g., \cite{chen2022focalclick,liu2023simpleclick}), and by adding lightweight high-resolution refinement heads for sharper boundaries \cite{kirillov2020pointrend}.
Foundation models further broaden promptability in 2D, most notably Segment Anything (SAM) \cite{kirillov2023segment} and its video extension \cite{ravi2024sam}.
Our work adapts the ``refine-where-needed'' principle to 3D by performing click-centric local refinement on a fine voxel grid and explicitly suppressing hard false positives.

\paragraph{3D point cloud segmentation.}
Backbones for 3D understanding span point-based networks \cite{qi2017pointnet,qi2017pointnet++,thomas2019kpconv,zhao2021point} and efficient large-scale designs \cite{hu2020randla}, as well as sparse convolutional approaches \cite{graham20183d,choy20194d,tang2020searching}.
For 3D instance segmentation, grouping-based methods such as PointGroup \cite{jiang2020pointgroup} and SoftGroup \cite{vu2022softgroup} are widely used,
while transformer-based mask prediction has shown strong performance in recent unified frameworks \cite{schult2023mask3d,kolodiazhnyi2024oneformer3d}.
NegROI is complementary to these backbones and focuses on interactive prompting, local refinement, and robust background modeling.

\paragraph{Interactive and promptable 3D segmentation.}
Interactive 3D segmentation has advanced from early click-based pipelines to transformer-driven promptable systems.
InterObject3D \cite{kontogianni2022interactive} and AGILE3D \cite{yue2023agile3d} demonstrate that attention-guided decoding can effectively integrate user clicks for object extraction in cluttered scenes.
Easy3D \cite{simonelli2025easy3d} provides a strong and simple interactive baseline, while Point-SAM \cite{zhou2024point} extends the SAM-style prompting paradigm to point clouds.
Compared to prior work, we explicitly model confusing background via scene-conditioned negative prompts and couple them with click-centric fine-grid refinement to improve boundary fidelity and cross-dataset robustness.

\paragraph{Open-vocabulary vision-language and 3D understanding.}
Open-vocabulary segmentation commonly builds on large-scale vision-language pretraining \cite{radford2021learning,jia2021scaling}.
Text-supervised grouping and segmentation emerge in models such as GroupViT \cite{xu2022groupvit}, and open-vocabulary recognition can be bootstrapped from image-level supervision \cite{zhou2022detecting} or open-set detectors \cite{liu2024grounding}.
For segmentation, CLIP-adapted mask models provide a practical recipe for open-vocabulary transfer \cite{liang2023open}.
In 3D, OpenScene \cite{peng2023openscene} studies open-vocabulary 3D scene understanding, while OpenMask3D \cite{takmaz2023openmask3d} and Open3DIS \cite{nguyen2024open3dis} address open-vocabulary 3D instance segmentation by leveraging 2D foundation models and mask guidance.
Our work targets the \emph{interactive} setting and is compatible with open-vocabulary querying, while focusing on improved error localization and suppression of hard distractors.

\paragraph{Hard negatives and boundary-focused refinement.}
A recurring challenge in interactive segmentation is that errors are highly concentrated around ambiguous boundaries and confusing distractors.
Prior 2D interactive methods improve click efficiency by explicitly focusing on hard regions \cite{chen2022focalclick,liu2023simpleclick} and by adding lightweight high-resolution heads to sharpen boundaries \cite{kirillov2020pointrend}.
In 3D interactive segmentation, clutter and object attachments further amplify boundary-proximal false positives \cite{kontogianni2022interactive,yue2023agile3d}.
Our approach follows the same philosophy of \emph{refine-where-needed} and \emph{focus-on-hard-negatives}, but instantiates it with (i) boundary-aware hard negative mining that supervises negative-prompt attention, and (ii) click-centric fine-grid ROI refinement that improves boundary fidelity without dense high-resolution decoding.

\section{Main Method}
\label{sec:method}

As aforementioned, we refer to our approach as \textit{NegROI}, which combines scene-conditioned negative prompts with click-centric ROI refinement for robust interactive 3D segmentation.

\subsection{Problem Statement}
Given a point cloud $\mathcal{P}=\{(\x_i,\mathbf{c}_i)\}_{i=1}^{P}$ with coordinates $\x_i\in\R^3$ and color/features $\mathbf{c}_i$,
we voxelize it into a coarse voxel set $\mathcal{V}=\{(\vcoord_j,\feat_j)\}_{j=1}^{V}$, where $\vcoord_j\in\Z^3$ and $\feat_j$ is a voxel feature.
In interactive segmentation, at step $c$ we are given clicks
$\mathcal{K}_c=\{(\kclick_t,y_t)\}_{t=1}^{c}$ with $y_t\in\{\text{pos},\text{neg}\}$.
We predict voxel logits $\slog^{(c)}\in\R^{V}$ and a binary mask by thresholding $\sigmoid(\slog^{(c)})$.

\paragraph{Click coordinates and voxel indexing.}
We denote each click location $\kclick_t$ in \emph{metric} coordinates, i.e., $\kclick_t\in\R^3$ in the same coordinate system as $\x_i$.
For operations on the coarse voxel grid (voxel size $v$), we quantize $\kclick_t$ to a voxel coordinate
$\tilde{\kclick}_t=\Pi_v(\kclick_t)\in\Z^3$ using the same voxelization rule as $\mathcal{V}$.
We further denote by $j_t\in\{1,\dots,V\}$ the index of the coarse voxel whose coordinate matches (or is nearest to) $\tilde{\kclick}_t$.
Unless otherwise specified, we use $\kclick_t$ for metric coordinates and $(\tilde{\kclick}_t,j_t)$ for coarse-voxel coordinates/indexing.

\begin{figure}[t]
    \centering
    \includegraphics[width=\textwidth]{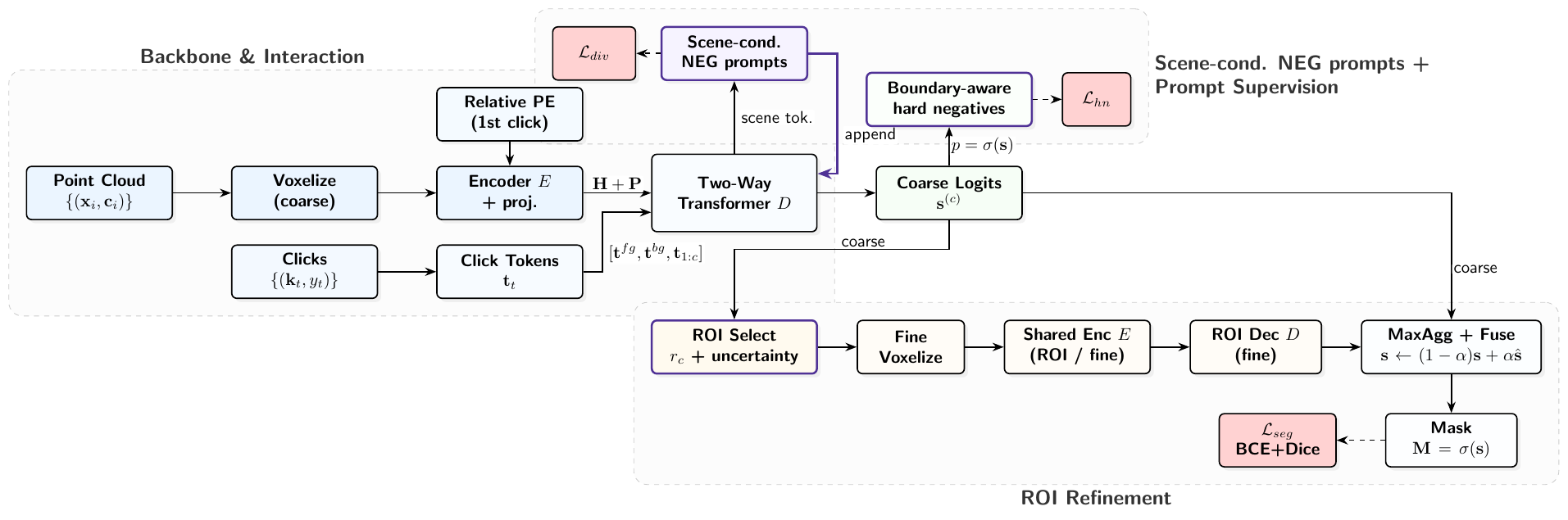}
    \caption{\textbf{NegROI overview (aligned with our implementation).} We use a sparse voxel backbone and the Easy3D two-way transformer decoder to produce coarse logits. We augment decoding with scene-conditioned negative prompts supervised by $\mathcal{L}_{hn}$ and $\mathcal{L}_{div}$. For boundary accuracy, we refine a click-centric ROI on a fine grid using the shared encoder and an ROI decoder, then max-aggregate and fuse back to coarse logits.}
    \label{fig:method_overview}
\end{figure}

\subsection{Framework Overview}
As illustrated in Fig.~\ref{fig:method_overview}, \textit{NegROI} consists of three coupled parts:
\textit{(i) Backbone \& interaction}, which encodes the coarse voxel grid and injects click tokens into a two-way transformer decoder;
\textit{(ii) Scene-conditioned negative prompts}, which explicitly model confusing background via $K$ negative prototypes and supervises their attention on boundary hard negatives; and
\textit{(iii) ROI refinement \& output}, which refines the prediction in a click-centered region on a fine grid and fuses it back to the coarse logits.

\subsection{Backbone and Interaction Tokens}

\paragraph{Coarse voxel encoding.}
A voxel encoder $E(\cdot)$ maps voxelized inputs to scene tokens as below:
\begin{equation}
\Hh = E(\mathcal{V}) \in \R^{V\times F}.
\end{equation}

\paragraph{Relative positional encoding.}
We use the quantized first click $\tilde{\kclick}_1$ as the coordinate origin on the coarse voxel grid.
For each voxel $j$,
\begin{equation}
\Delta \vcoord_j = \vcoord_j - \tilde{\kclick}_1, \qquad
\Pp_j = W_p\,\phi\!\left(\Delta\vcoord_j / S\right)\in\R^{F},
\end{equation}
where $\phi(\cdot)$ is harmonic encoding. We set $S$ to the per-scene voxel-grid scale, e.g.,
\begin{equation}
S=\max\{(v_{\max}-v_{\min})_x,\,(v_{\max}-v_{\min})_y,\,(v_{\max}-v_{\min})_z\},
\end{equation}
with $v_{\min},v_{\max}\in\Z^3$ being the min/max voxel coordinates of the scene.

\paragraph{Click tokens.}
We construct interaction tokens consisting of two learnable mask tokens $\mathbf{t}^{fg},\mathbf{t}^{bg}\in\R^F$ and $c$ click tokens.
For click $t$, we map it to the coarse voxel index $j_t$ (defined in Sec.~\ref{sec:method}) and form
\begin{equation}
\mathbf{t}_t = \Pp_{j_t} + \mathbf{e}(y_t),
\end{equation}
where $\mathbf{e}(\cdot)$ is a learned embedding for click type.
The base token sequence is
\begin{equation}
\Tt^{(c)}_{\text{base}} = \left[\mathbf{t}^{fg},\mathbf{t}^{bg},\mathbf{t}_1,\ldots,\mathbf{t}_c \right]\in\R^{(2+c)\times F}.
\end{equation}

\subsection{Two-Way Transformer Decoder}
We adopt the two-way transformer decoder from Easy3D \cite{simonelli2025easy3d}, which exchanges information between scene tokens and interaction tokens:
\begin{equation}
(\tilde{\Hh},\tilde{\Tt}) = D(\Hh,\Pp,\Tt).
\end{equation}
Coarse voxel logits are obtained via a foreground/background token difference:
\begin{equation}
s^{(c)}_j = \langle \tilde{\mathbf{t}}^{fg}, \tilde{\mathbf{h}}_j \rangle - \langle \tilde{\mathbf{t}}^{bg}, \tilde{\mathbf{h}}_j \rangle.
\label{eq:coarse_logits}
\end{equation}

\subsection{Scene-Conditioned Negative Prompts and Supervision}
\label{sec:negprompts}
Hard false positives often originate from confusing background structures.
To explicitly model such distractors, we introduce $K$ learnable negative prompt queries $\{\mathbf{q}_k\}_{k=1}^{K}$ and compute scene-conditioned prototypes by cross-attending to the scene tokens:
\begin{equation}
\mathbf{p}^{(c)}_k = \mathrm{Attn}\!\left(\mathbf{q}_k,\; \Hh+\Pp,\; \Hh+\Pp\right)\in\R^F,
\qquad
\hat{\mathbf{p}}^{(c)}_k = \mathrm{MLP}(\mathbf{p}^{(c)}_k) + \mathbf{e}(\text{neg}).
\end{equation}
We append $\{\hat{\mathbf{p}}^{(c)}_k\}$ to the interaction tokens before decoding:
\begin{equation}
\Tt^{(c)} = \left[\Tt^{(c)}_{\text{base}}, \hat{\mathbf{p}}^{(c)}_1,\ldots,\hat{\mathbf{p}}^{(c)}_K \right].
\end{equation}

\paragraph{Prompt diversity regularization.}
To avoid prompt collapse, we encourage diversity among negative prototypes at each click step by penalizing off-diagonal cosine similarity:
\begin{equation}
\mathcal{L}_{\text{div}}^{(c)}
=
\frac{1}{K(K-1)}
\sum_{i\neq j}
\left(\cos(\mathbf{p}^{(c)}_{i},\mathbf{p}^{(c)}_{j})\right)^2.
\label{eq:div}
\end{equation}
where $\{\mathbf{p}^{(c)}_{k}\}_{k=1}^{K}$ are the scene-conditioned negative prototypes computed at click step $c$.

\paragraph{Boundary-aware hard negative supervision.}
Let $A^{(c)}\in\R^{K\times V}$ denote the attention weights produced when generating negative prototypes (averaged across heads), and $\bar{A}^{(c)}=\frac{1}{K}\sum_k A^{(c)}_k$.
We supervise $\bar{A}^{(c)}$ to focus on hard background near object boundaries.
Specifically, we define boundary-adjacent background candidates
\begin{equation}
\mathcal{B} = \left\{j \mid y_j=0,\;\exists j'\in\mathcal{N}(j)\ \text{s.t.}\ y_{j'}=1 \right\},
\end{equation}
where $\mathcal{N}(j)$ is the 6-neighborhood in the voxel grid and $y_j$ denotes the ground-truth voxel label.
Among candidates in $\mathcal{B}$ (falling back to all background if $\mathcal{B}$ is too small), we select the top-$k$ voxels with highest predicted foreground probability $p_j=\sigmoid(s^{(c)}_j)$ to form $\mathcal{H}$ and define a uniform target distribution
$t_j=\frac{1}{|\mathcal{H}|}\mathbf{1}[j\in\mathcal{H}]$.
We minimize cross-entropy:
\begin{equation}
\mathcal{L}_{\text{hn}}^{(c)} = -\sum_{j=1}^{V} t_j \log\left(\bar{A}^{(c)}_j+\epsilon \right).
\label{eq:hn}
\end{equation}

\subsection{Click-Centric ROI Refinement and Output}
\label{sec:roi_refine}
Coarse voxels may blur thin structures and boundaries.
We therefore refine locally around the \emph{current click} on a finer grid and fuse the refined logits back to the coarse prediction, as shown in Fig.~\ref{fig:method_overview}.

\paragraph{ROI selection with adaptive radius.}
Let the current click center be $\kclick_c$ (metric coordinates). We select ROI points
\begin{equation}
\mathcal{P}_{roi}=\{ i \mid \|\x_i-\kclick_c\|\le r_c \},
\end{equation}
where $r_c$ is an adaptive radius predicted from click context and local density (Sec.~\ref{sec:radius}).
We voxelize $\mathcal{P}_{roi}$ at a finer resolution $v_f = v/\eta$ (fine scale $\eta$) to obtain fine voxels $\mathcal{V}_f$ and fine scene tokens $\Hh_f\in\R^{V_f\times F}$.

\paragraph{Shared encoder and ROI decoder.}
We reuse the same encoder $E$ to embed the fine ROI voxels (weight sharing with the coarse branch).
We then apply the same Easy3D-style two-way decoder $D$ to the ROI tokens (again weight-shared), producing fine logits $\slog_f^{(c)}$.
Inside the ROI branch, we optionally build \emph{ROI-conditioned} negative prompts by cross-attending learnable NEG queries to the ROI tokens, and append them to the ROI interaction tokens (``ROI-cond.\ NEG'' in Fig.~\ref{fig:method_overview}).

\paragraph{Fine-to-coarse max aggregation and fusion.}
Each fine voxel maps to a coarse voxel by $\lfloor \mathbf{u}/\eta \rfloor$. We aggregate by max:
\begin{equation}
\hat{s}^{(c)}_j = \max_{u\in\mathcal{M}(j)} s^{(c)}_{f,u},
\end{equation}
and fuse with the coarse logits using a residual update:
\begin{equation}
s^{(c)}_j \leftarrow (1-\alpha)\,s^{(c)}_j + \alpha\,\hat{s}^{(c)}_j.
\label{eq:fuse}
\end{equation}

\subsubsection{Adaptive Radius Prediction.}
\label{sec:radius}
We begin with a decayed base radius $r^{\text{base}}_c = r_0\,\gamma^{c-1}$.
We estimate local density around the click using $k$NN distances and compute mean/median $(\mu,m)$.
We predict a log-scale adjustment
\begin{equation}
\Delta_c = \mathrm{MLP}\!\left([\mathbf{t}_c,\log \mu,\log m,\log r^{\text{base}}_c, c/C]\right),
\qquad
r_c = \mathrm{clip}\!\left(r^{\text{base}}_c \exp(\Delta_c)\right),
\label{eq:radius}
\end{equation}
where $\mathrm{clip}$ clamps to $[r_{\min},r_{\max}]$.
If the ROI contains too few points, we enlarge $r_c$ (bounded) a few times to ensure stable refinement.

\subsubsection{Uncertainty-Driven Selective Refinement.}
We prioritize ambiguous regions inside the ROI using coarse uncertainty:
\begin{equation}
u_j = 1 - 2\left|\sigmoid(s^{(c)}_j)-0.5\right| \in [0,1].
\label{eq:uncert}
\end{equation}
Each ROI point is mapped to its coarse voxel index $v(i)$; we keep points with $u_{v(i)}\ge\tau$ (fallback to the unfiltered ROI if too few points remain), focusing fine decoding on boundary and hard regions.

\subsection{Training Objective}
At click step $c$, we use the segmentation loss
\begin{equation}
\mathcal{L}_{\text{seg}}^{(c)} =
\mathcal{L}_{\text{bce}}(\slog^{(c)},\mathbf{y})
+
\mathcal{L}_{\text{dice}}(\slog^{(c)},\mathbf{y}),
\end{equation}
and add scene-conditioned negative prompts regularizers:
\begin{equation}
\mathcal{L}
=
\frac{1}{C}\sum_{c=1}^{C}
\left(
\mathcal{L}_{\text{seg}}^{(c)}
+
\lambda_{\text{hn}}\mathcal{L}_{\text{hn}}^{(c)}
+
\lambda_{\text{div}}\mathcal{L}_{\text{div}}^{(c)}
\right).
\label{eq:final_loss}
\end{equation}

\paragraph{Click simulation.}
During training, we simulate interactions using an oracle policy that places the next click at informative error regions between prediction and ground truth.

\paragraph{Implementation note (DDP stability).}
In our implementation, adaptive radii affect ROI selection via a hard threshold.
To avoid unused-parameter issues in DDP, we optionally include a tiny regularizer on $\Delta_c$ with detached inputs:
\begin{equation}
\mathcal{L}_{\text{rad}} = \lambda_{\text{rad}} \|\Delta_c\|_2^2.
\end{equation}

% ============================================================
\section{Experiments}
\label{sec:experiments}

In this section, we conducted extensive experiments on several common benchmarks to empirically evaluate the advantages of our method by comparing with the leading interactive and non-interactive baselines. We also did ablation studies to show the effectiveness of our every modules.

\subsection{Benchmarks}
We evaluate interactive 3D instance segmentation on three widely-used benchmarks:
\textbf{ScanNet40} \cite{dai2017scannet}, \textbf{S3DIS} \cite{armeni20163d}, and \textbf{KITTI-360} \cite{liao2022kitti}.
ScanNet40 and S3DIS represent cluttered indoor environments with frequent object attachments and ambiguous boundaries,
while KITTI-360 features large-scale outdoor scans with sparse geometry and long-range context.
Unless otherwise stated, we follow the official dataset splits and evaluation protocols adopted in prior interactive 3D segmentation works
(e.g., \cite{kontogianni2022interactive,yue2023agile3d}).

\subsection{Baselines}
We compare against representative interactive 3D segmentation baselines, including click-conditioned transformer decoders without ROI refinement,
multi-resolution refinement methods without explicit background prompting, and prompt/query-based interactive approaches
(e.g., \cite{kontogianni2022interactive,yue2023agile3d,zhou2024point}).
When official implementations are available, we run them under the same evaluation protocol.
Otherwise, we follow the paper descriptions and ensure identical click simulation, data preprocessing, and evaluation metrics.
We also report Easy3D\cite{simonelli2025easy3d} as a strong promptable interactive baseline in our setting.

\subsection{Interactive Evaluation Protocol}
\paragraph{Click simulation.}
To ensure fair and repeatable evaluation, we adopt a standard simulated-click protocol.
For each target instance, the first click is sampled inside the ground-truth region.
At each subsequent step, we update the mask prediction and place the next click on the largest error region:
a positive click on the largest false-negative component if under-segmentation dominates, otherwise a negative click on the largest false-positive component.
This policy approximates an oracle user that always clicks the most informative mistake and is commonly used for benchmarking interactive methods.

\paragraph{Target specification (prompts).}
Our primary setting is class-agnostic interactive instance segmentation, where the target instance is specified by clicks only.
If a baseline requires an additional query signal (e.g., an open-vocabulary text prompt), we use simple class-name prompts with a consistent template across methods.

\paragraph{Metrics.}
We report point-wise IoU (\%) under click budgets $k\in\{1,2,3,5,10\}$, denoted as IoU@k.
% , and plot click-efficiency curves (IoU vs.\ number of clicks).
% When relevant, we also report voxel-level IoU and NoC@$\tau$ (the number of clicks required to reach an IoU threshold $\tau$).
All methods are evaluated with the same click simulator and the same preprocessing on each dataset to isolate algorithmic differences.

\subsection{Implementation Details}
\paragraph{Backbone and voxelization.}
Our model uses a sparse voxel encoder and a two-way transformer decoder.
We follow dataset-standard sparse voxel preprocessing and keep voxelization consistent between training and testing.
To improve boundary accuracy without incurring dense high-resolution costs, our refinement branch re-voxelizes only within selected ROIs using a finer resolution than the coarse grid.

\paragraph{Scene-conditioned negative prompts and click-centric ROI refinement.}
NegROI combines \emph{scene-conditioned negative prompts} with \emph{click-centric ROI refinement} (Fig.~\ref{fig:method_overview}).
At each click step, we first decode coarse mask logits using click tokens augmented by $K$ scene-conditioned negative prototypes.
We then refine boundaries with a click-centric ROI branch: we select an ROI centered at the current click using an adaptive radius $r_c$ and a coarse uncertainty signal,
re-voxelize the ROI on a finer grid, and run the \emph{shared encoder} $E$ together with an \emph{ROI decoder} to produce fine logits.
Finally, we \emph{max-aggregate} fine logits to the coarse grid and \emph{fuse} them back with weight $\alpha$.
For prompt learning, we supervise negative-prompt attention via a boundary-aware hard-negative loss $\mathcal{L}_{hn}$ and regularize prompt collapse with the diversity loss $\mathcal{L}_{div}$.
We tune these hyperparameters on a ScanNet40 validation split and keep them fixed across datasets.

\paragraph{Hyperparameters.}
Unless otherwise stated, we use $K{=}8$ negative prompts, fine scale $\eta{=}2$ (i.e., $v_f{=}v/\eta$), uncertainty threshold $\tau{=}0.20$, and fusion weight $\alpha{=}0.7$.
For the click-centric ROI, we set the base radius $r_0{=}1.0$m with decay $\gamma{=}0.9$, and clamp the adaptive radius to $[r_{\min},r_{\max}]{=}[0.35r_0,\,3.0r_0]$.
We estimate local density with $k$NN using $k{=}32$, and mine hard negatives with top-$k$ background voxels using $k{=}256$.
All these hyperparameters are tuned once on a ScanNet40 validation split and then fixed for all datasets.

\paragraph{Training.}
We train end-to-end using the segmentation loss on the fused logits and auxiliary objectives:
$\mathcal{L}_{seg}=\mathcal{L}_{bce}+\mathcal{L}_{dice}$, plus the prompt regularizers $\lambda_{hn}\mathcal{L}_{hn}$ and $\lambda_{div}\mathcal{L}_{div}$.
We simulate up to $C$ clicks per query during training using the oracle policy described above.

\subsection{Main Results}
Table~\ref{tab:openvocab3d} reports the main comparison under a unified \emph{train-on-ScanNet40} setting.
We train all methods on ScanNet40 and evaluate on three test benchmarks:
\textbf{ScanNet40} (in-domain) and \textbf{S3DIS}/\textbf{KITTI-360} (out-of-domain).
This protocol directly measures cross-dataset robustness under substantial domain shifts in scene scale, point density, and sensor characteristics.

\paragraph{In-domain performance.}
On ScanNet40, NegROI consistently improves IoU across click budgets, with the largest gains under sparse interactions (IoU@1--IoU@3).
These improvements align with our design goals: scene-conditioned negative prompts reduce hard distractors and attached clutter early,
while the click-centric fine-grid ROI branch sharpens boundaries through re-voxelization and ROI decoding, followed by max-aggregation and fusion back to coarse logits.

\paragraph{Out-of-domain generalization.}
When transferring to S3DIS and KITTI-360 without any target-domain training, NegROI maintains strong performance and outperforms prior baselines across most click budgets.
The gains are particularly pronounced at low click counts, where boundary ambiguity and hard distractors dominate.
This behavior is consistent with (i) boundary-aware hard-negative prompt supervision ($\mathcal{L}_{hn}$), which suppresses boundary-proximal false positives, and
(ii) click-centric ROI refinement, which allocates fine-grained computation to the most informative region indicated by the click.

\begin{table}[t]
\centering
\caption{\textbf{Main results under a unified train-on-ScanNet40 protocol.}
All methods are trained on ScanNet40 and tested on ScanNet40 (in-domain), S3DIS and KITTI-360 (out-of-domain).
Numbers denote point-wise IoU (\%) under click budgets $k\in\{1,2,3,5,10\}$ (IoU@k). Best results are in bold.}
\label{tab:openvocab3d}
\setlength{\tabcolsep}{6pt}
\renewcommand{\arraystretch}{1.5}
\begin{tabular}{l l c c c c c}
\toprule
{\bf Dataset} & {\bf Method} & IoU@1 & IoU@2 & IoU@3 & IoU@5 & IoU@10 \\
\midrule
\multirow{4}{*}{ScanNet40} &
InterObject3D\cite{kontogianni2022interactive}  & 40.8 & 55.9 & 63.9 & 67.6 & 77.6 \\
& AGILE3D\cite{yue2023agile3d}       & 63.0 & 70.6 & 75.1 & 79.7 & 83.5 \\
& Easy3D\cite{simonelli2025easy3d}            & 68.2 & 74.6 & 77.3 & 79.6 & 81.7 \\
& {\bf NegROI (Ours)}              & \textbf{72.1} & \textbf{78.1} & \textbf{80.9} & \textbf{82.3} & \textbf{83.6} \\
\midrule
\multirow{5}{*}{S3DIS} &
InterObject3D\cite{kontogianni2022interactive}  & 38.5 & 54.0 & 62.5 & 72.4 & 79.9 \\
& AGILE3D\cite{yue2023agile3d}        & 58.5 & 70.7 & 77.4 & 83.6 & \textbf{88.3} \\
& Point-SAM\cite{zhou2024point}     & 38.8 & n/a  & 67.1 & 72.2 & 80.6 \\
& Easy3D\cite{simonelli2025easy3d}            & 65.7 & 76.0 & 80.8 & 84.9 & 87.8 \\
& {\bf NegROI (Ours)}             & \textbf{66.9} & \textbf{77.8} & \textbf{82.2} & \textbf{85.4} & 87.8 \\
\midrule
\multirow{5}{*}{KITTI-360} &
InterObject3D\cite{kontogianni2022interactive} & 2.0  & 5.1  & 8.5  & 72.4 & 83.6 \\
& AGILE3D\cite{yue2023agile3d}        & 34.8 & 40.7 & 42.7 & 44.4 & 49.6 \\
& Point-SAM\cite{zhou2024point}     & 44.0 & n/a  & 67.1 & 72.2 & 80.8 \\
& Easy3D\cite{simonelli2025easy3d}            & 46.3 & 58.7 & 66.7 & 76.2 & 83.6 \\
& {\bf NegROI (Ours)}              & \textbf{47.5} & \textbf{62.6} & \textbf{71.4} & \textbf{80.1} & \textbf{87.7} \\
\bottomrule
\end{tabular}
\end{table}

\subsection{Comparison with Non-Interactive Methods}
Although NegROI is designed for interactive refinement, we further evaluate its \emph{single-click} performance under the ScanNet20 protocol commonly used for assessing generalization to unseen categories.
Specifically, we train models \emph{only} on ScanNet20 and test in two settings:
ScanNet20 $\rightarrow$ ScanNet20 (only seen objects) and ScanNet20 $\rightarrow$ ScanNet40 (seen + unseen objects).
Table~\ref{tab:non_interactive} compares NegROI with representative non-interactive/open-vocabulary 3D instance segmentation baselines, including Mask3D~\cite{schult2023mask3d}, AGILE3D~\cite{yue2023agile3d}, and Easy3D~\cite{simonelli2025easy3d}.
NegROI improves mAP as well as AP$_{50}$/AP$_{25}$ in both the in-domain (seen-only) and the seen$\,+\,$unseen transfer setting, indicating that our scene-conditioned negative prompting and click-centric refinement also benefit early (single-click) predictions and enhance robustness beyond the ScanNet40-trained interactive evaluation in Table~\ref{tab:openvocab3d}.

\begin{table}[t]
\centering
\caption{\textbf{Single-click evaluation for models trained only on ScanNet20.}
We report results under the ScanNet20 $\rightarrow$ ScanNet20 setting (only seen objects) and the ScanNet20 $\rightarrow$ ScanNet40 setting (seen + unseen objects).}
\label{tab:non_interactive}
\setlength{\tabcolsep}{10pt}
\renewcommand{\arraystretch}{1.5}
\begin{tabular}{l l c c c}
\toprule
\textbf{Setting} & \textbf{Method} & {mAP}  & ${AP_{50}}$  & ${AP_{25}}$  \\
\midrule
\multirow{4}{*}{ScanNet20 $\rightarrow$ ScanNet20}
& Mask3D~\cite{schult2023mask3d} & 51.5 & 77.0 & 90.2 \\
& AGILE3D~\cite{yue2023agile3d} & 53.5 & 75.6 & 91.3 \\
& Easy3D~\cite{simonelli2025easy3d} & 56.1 & 79.5 & 93.1 \\
& {\bf NegROI (Ours)} & \textbf{59.6} & \textbf{82.3} & \textbf{95.2} \\
\midrule
\multirow{4}{*}{ScanNet20 $\rightarrow$ ScanNet40}
& Mask3D~\cite{schult2023mask3d} & 5.3 & 13.1 & 24.7 \\
& AGILE3D~\cite{yue2023agile3d} & 24.8 & 45.7 & 72.4 \\
& Easy3D~\cite{simonelli2025easy3d} & 39.2 & 64.6 & 85.5 \\
& {\bf NegROI (Ours)} & \textbf{43.0} & \textbf{67.9} & \textbf{87.8} \\
\bottomrule
\end{tabular}
\end{table}

\subsection{Ablation Studies}
We conduct ablations to quantify the contribution of each component and the proposed regularizers.

\paragraph{Component ablation.}
We progressively enable (i) scene-conditioned negative prompts, (ii) click-centric fine-grid ROI refinement, (iii) uncertainty-driven ROI selection,
(iv) boundary-aware hard negative supervision for prompt attention ($\mathcal{L}_{hn}$), and (v) prompt diversity regularization ($\mathcal{L}_{div}$).
Table~\ref{tab:ablation} summarizes the results on ScanNet40 and S3DIS.

\begin{table}[t]
\centering
\caption{\textbf{Component ablation on ScanNet40 and S3DIS.} Numbers denote point-wise IoU (\%) under click budgets $k\in\{1,5,10\}$.}
\label{tab:ablation}
\renewcommand{\arraystretch}{1.5}
\begin{tabular}{l c c c @{\hspace{6pt}} c c c}
\toprule
\multirow{2}{*}{\bf Variant} &
\multicolumn{3}{c}{\bf ScanNet40} & \multicolumn{3}{c}{\bf S3DIS} \\
\cmidrule(lr){2-4}\cmidrule(lr){5-7}
& IoU@1 & IoU@5 & IoU@10 & IoU@1 & IoU@5 & IoU@10 \\
\midrule
Base (no ROI, no NEG prompts)                  & 67.0 & 78.8 & 80.8 & 62.5 & 80.8 & 83.8 \\
+ Scene-cond. NEG prompts                       & 69.2 & 80.4 & 82.3 & 64.2 & 83.0 & 85.7 \\
+ ROI refinement (fine-grid)                    & 70.9 & 81.6 & 83.0 & 65.8 & 84.1 & 86.9 \\
+ Uncertainty-driven ROI selection              & 71.8 & 81.9 & 82.8 & 66.5 & 85.0 & 87.4 \\
+ Boundary hard negatives ($\mathcal{L}_{hn}$)  & 71.2 & 82.1 & 83.1 & 65.6 & 85.2 & 87.1 \\
+ Diversity ($\mathcal{L}_{div}$) (full)        & \textbf{72.1} & \textbf{82.3} & \textbf{83.6} & \textbf{66.9} & \textbf{85.4} & \textbf{87.8} \\
\bottomrule
\end{tabular}
\end{table}

Scene-conditioned negative prompts boost performance under sparse clicks (e.g., IoU@1: 67.0$\rightarrow$69.2 on ScanNet40 and 62.5$\rightarrow$64.2 on S3DIS).
Adding click-centric fine-grid ROI refinement further improves boundary quality (e.g., ScanNet40 IoU@5: 80.4$\rightarrow$81.6).
Uncertainty-driven ROI selection brings additional gains (e.g., ScanNet40 IoU@1: 70.9$\rightarrow$71.8).
Overall, combining the proposed regularizers yields the best full model on both datasets (ScanNet40: 72.1/82.3/83.6; S3DIS: 66.9/85.4/87.8 for IoU@1/5/10).

\paragraph{Loss ablation.}
We further examine the effect of the proposed prompt regularizers.
As shown in Table~\ref{tab:ablation}, adding boundary-aware hard negatives ($\mathcal{L}_{hn}$) and prompt diversity ($\mathcal{L}_{div}$) improves the overall results, with the full model achieving the best performance across datasets and click budgets.
In particular, $\mathcal{L}_{div}$ provides consistent gains when combined with $\mathcal{L}_{hn}$, suggesting that preventing prompt collapse helps stabilize negative guidance during refinement.

\subsection{Qualitative Results}
\begin{figure}[t]
\centering
\setlength{\tabcolsep}{1pt}
\renewcommand{\arraystretch}{1.0}

\newcommand{\imgw}{0.130\linewidth}
\newcommand{\dscol}{7mm}
\newcommand{\DS}[1]{\rotatebox{90}{\scriptsize\textbf{#1}}}
\newcommand{\tbd}{\textit{--}}

% ---- Cell macros (Writing style 1) ----
% #1: image path, #2: text below image
\newcommand{\QualCellImg}[2]{%
  \makecell[c]{\includegraphics[width=\imgw]{#1}\\[-1.5pt]\scriptsize #2}%
}

% #1: image path, no visible text (keeps same height via phantom)
\newcommand{\QualCellImgNoTxt}[1]{%
  \makecell[c]{\includegraphics[width=\imgw]{#1}\\[-1.5pt]\scriptsize\phantom{IoU@1 = 00.0}}%
}

\begin{tabular}{@{}>{\centering\arraybackslash}m{\dscol} c c c c c c c@{}}
\toprule
& \small\textbf{Target}
& \multicolumn{3}{c}{\footnotesize\textbf{Easy3D}}
& \multicolumn{3}{c}{\footnotesize\textbf{NegROI (Ours)}}\\
\cmidrule(lr){2-2}\cmidrule(lr){3-5}\cmidrule(lr){6-8}

% ---------------- ScanNet40 ----------------
\DS{ScanNet40}
& \QualCellImgNoTxt{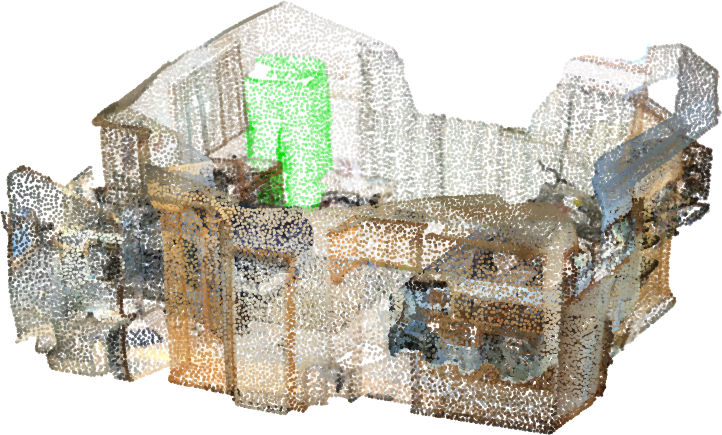}
& \QualCellImg{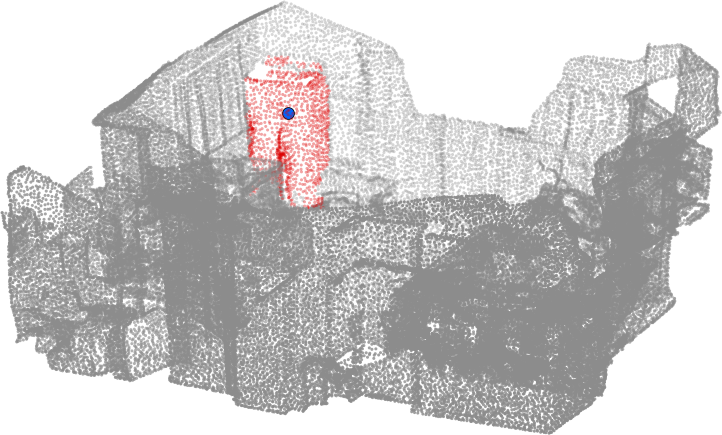}{IoU@1=68.8}
& \QualCellImg{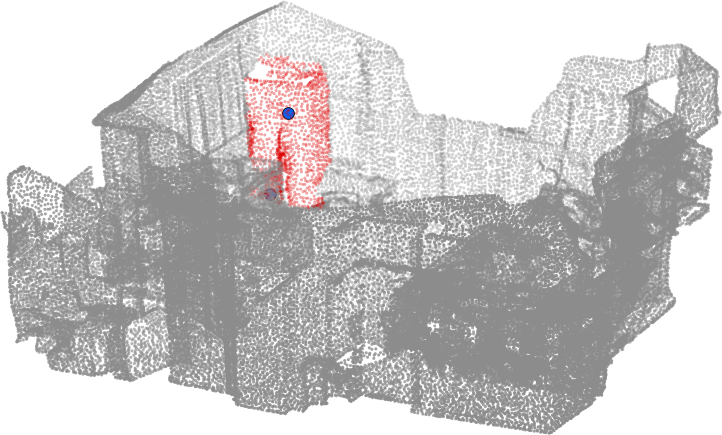}{IoU@2=82.5}
& \QualCellImg{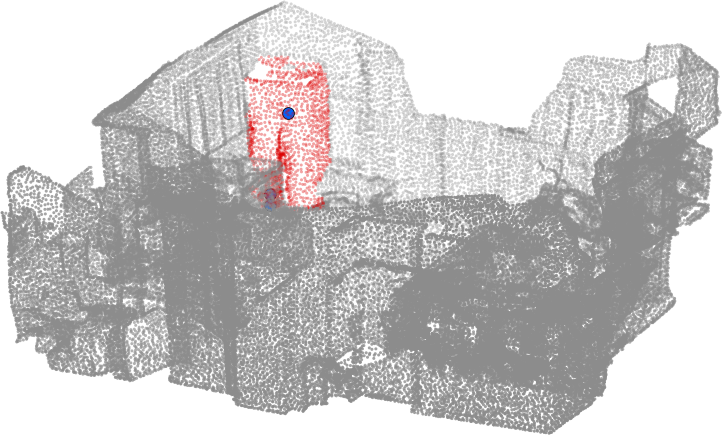}{IoU@3=85.9}
& \QualCellImg{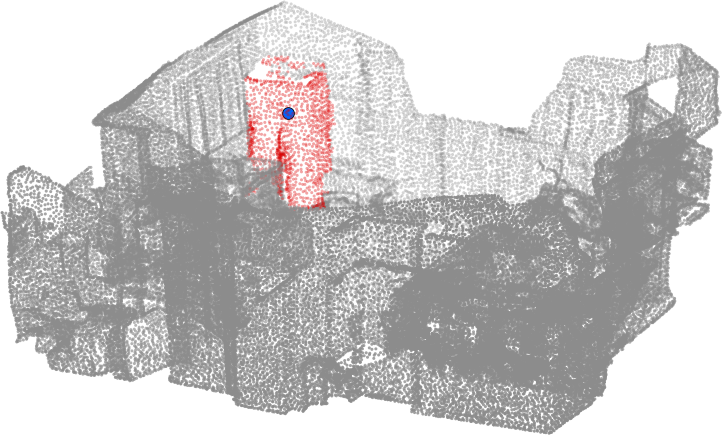}{IoU@1=70.8}
& \QualCellImg{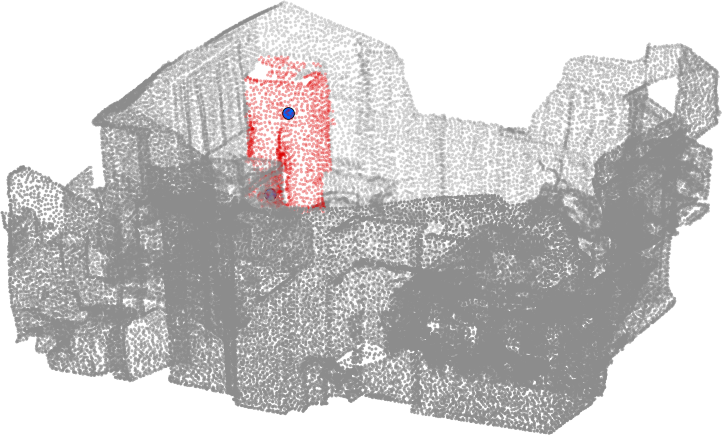}{IoU@2=86.0}
& \QualCellImg{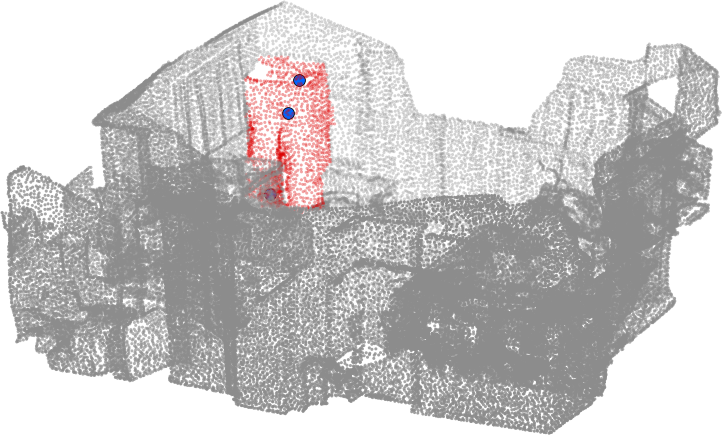}{IoU@3=88.0}\\
\addlinespace[3pt]

% ---------------- S3DIS ----------------
\DS{S3DIS}
& \QualCellImgNoTxt{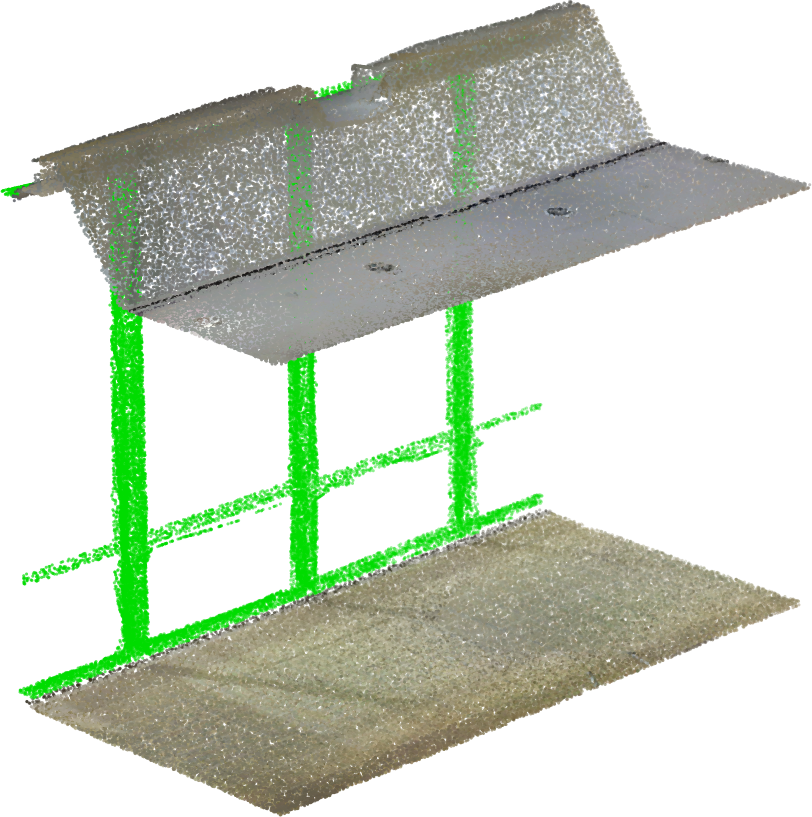}
& \QualCellImg{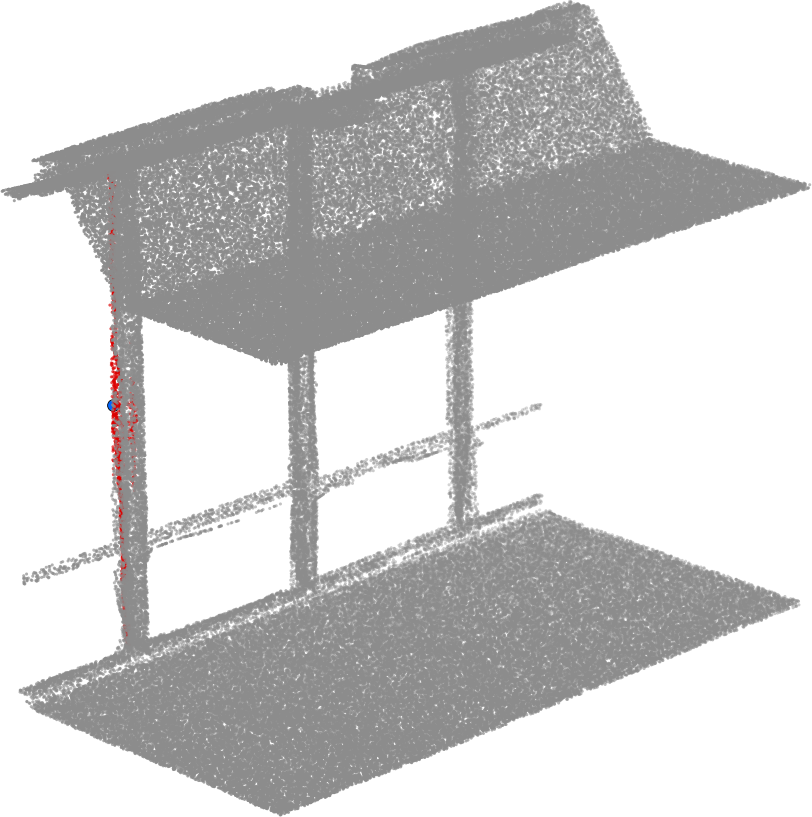}{IoU@1=5.1}
& \QualCellImg{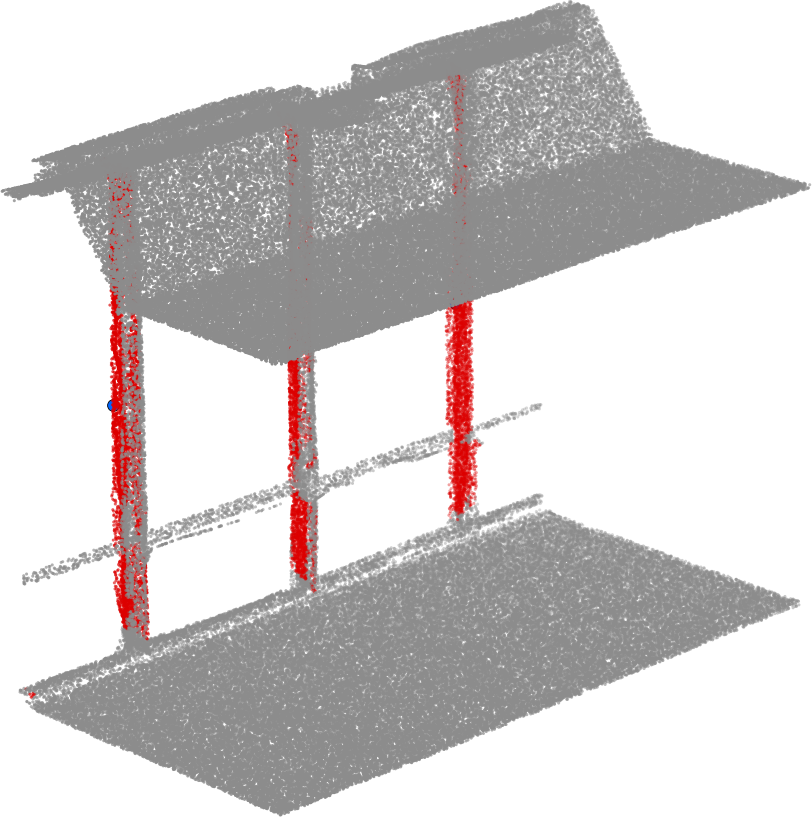}{IoU@2=43.2}
& \QualCellImg{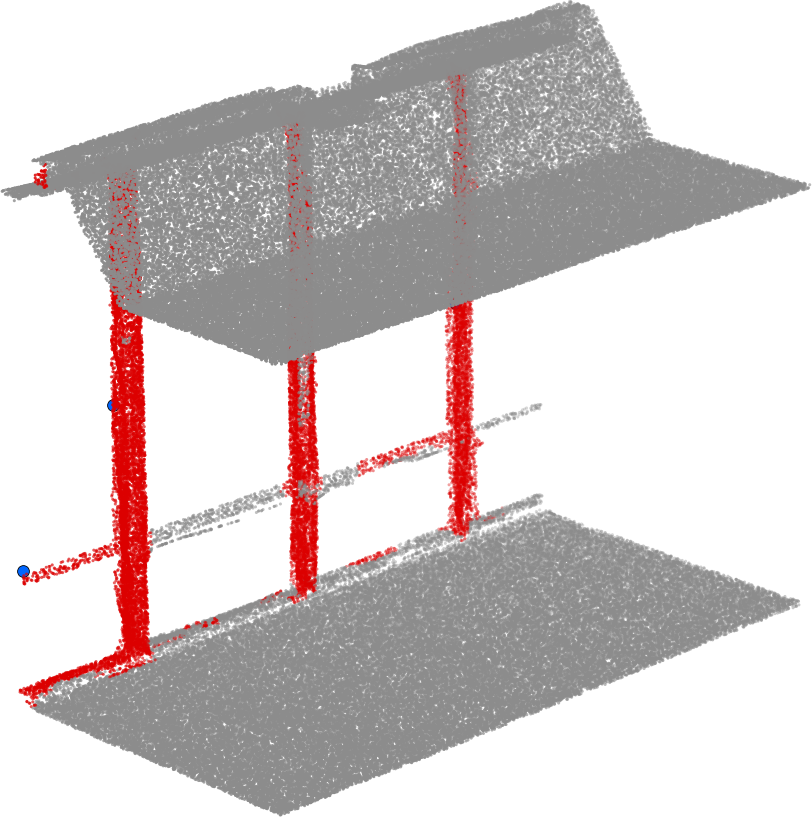}{IoU@3=69.5}
& \QualCellImg{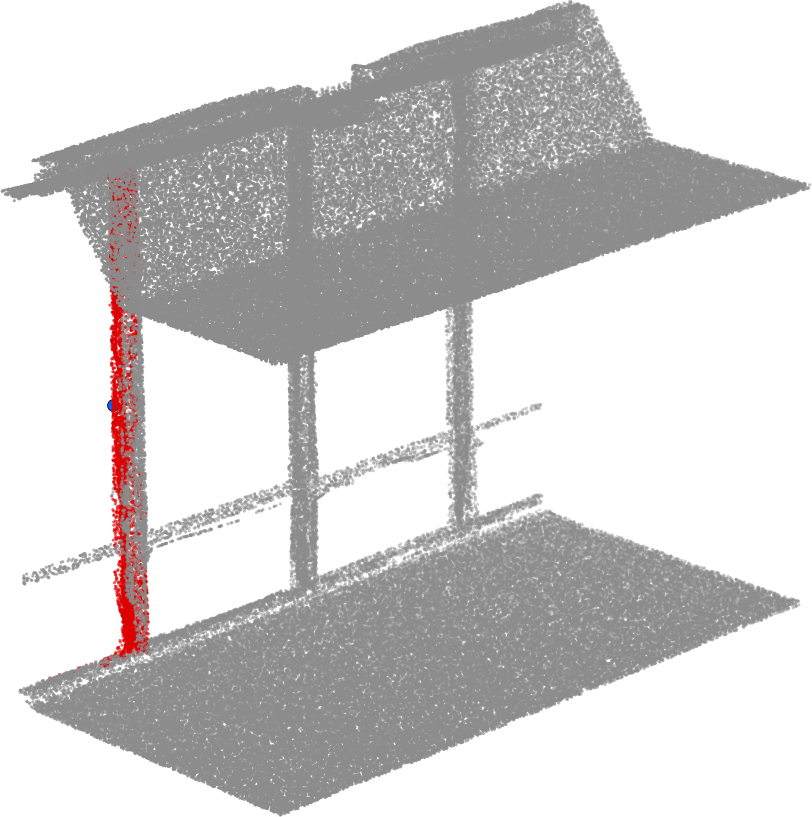}{IoU@1=18.6}
& \QualCellImg{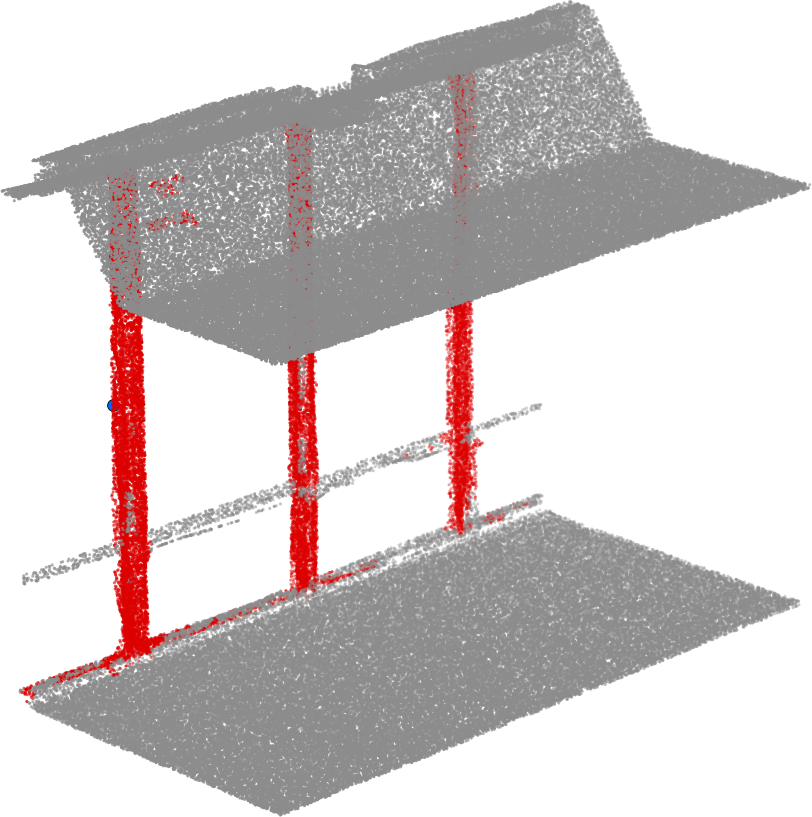}{IoU@2=70.5}
& \QualCellImg{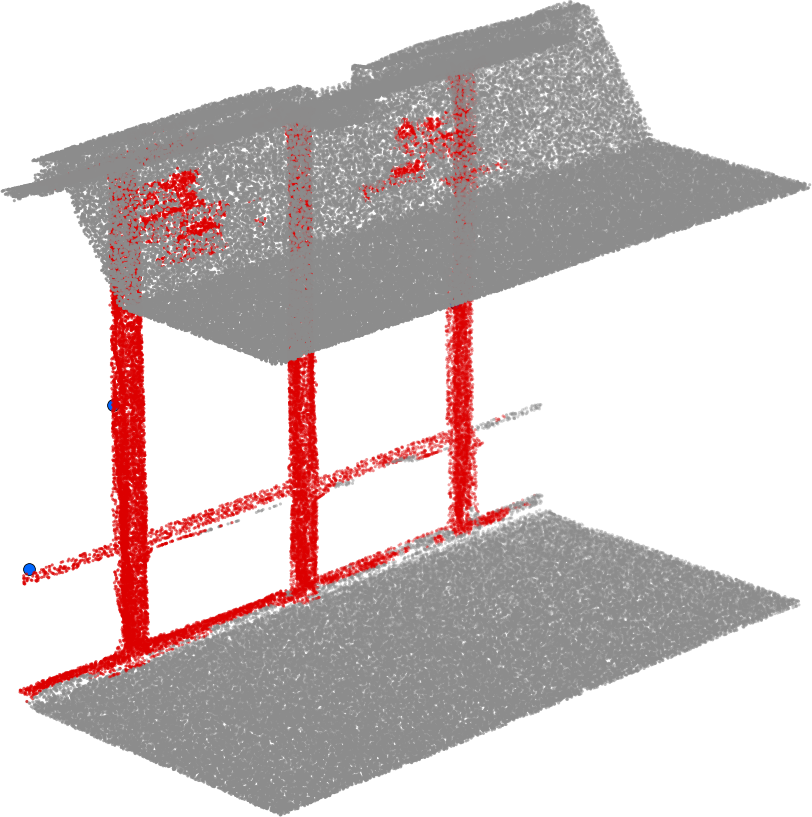}{IoU@3=76.8}\\
\addlinespace[3pt]

% ---------------- KITTI-360 ----------------
\DS{KITTI-360}
& \QualCellImgNoTxt{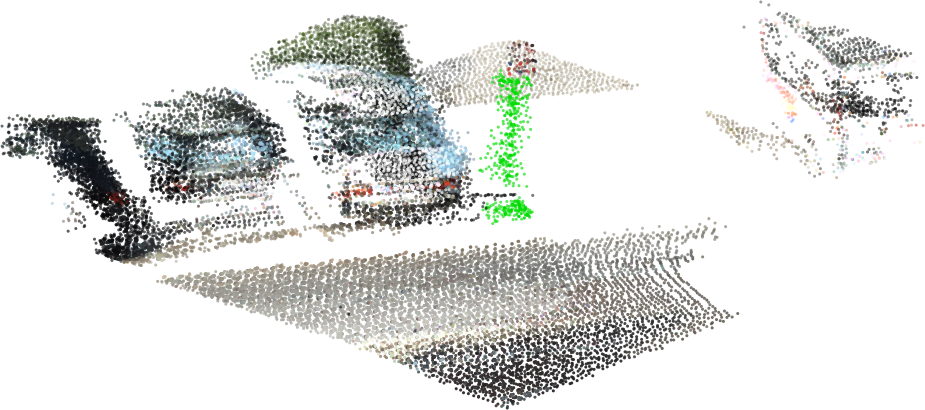}
& \QualCellImg{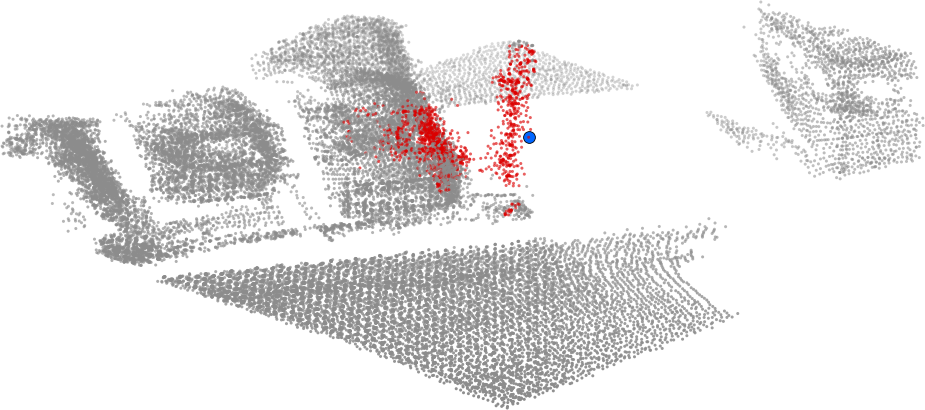}{IoU@1=25.2}
& \QualCellImg{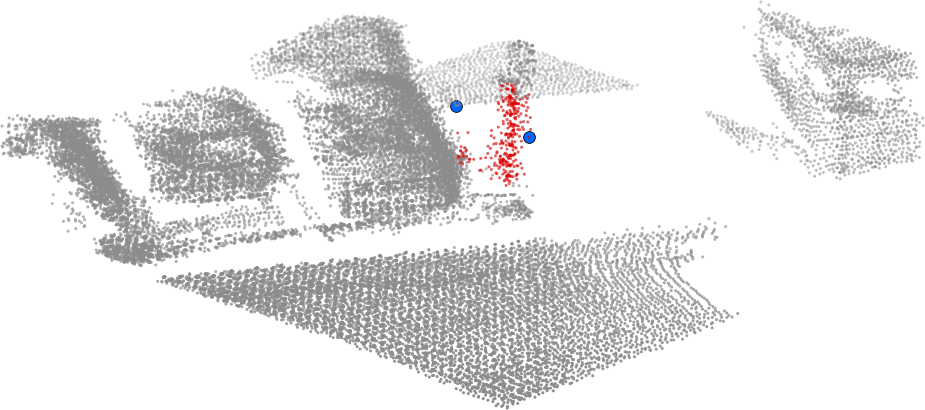}{IoU@2=51.3}
& \QualCellImg{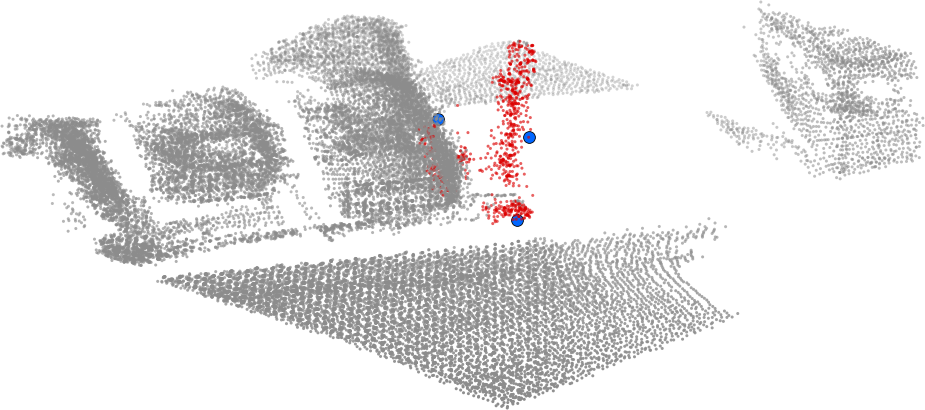}{IoU@3=67.6}
& \QualCellImg{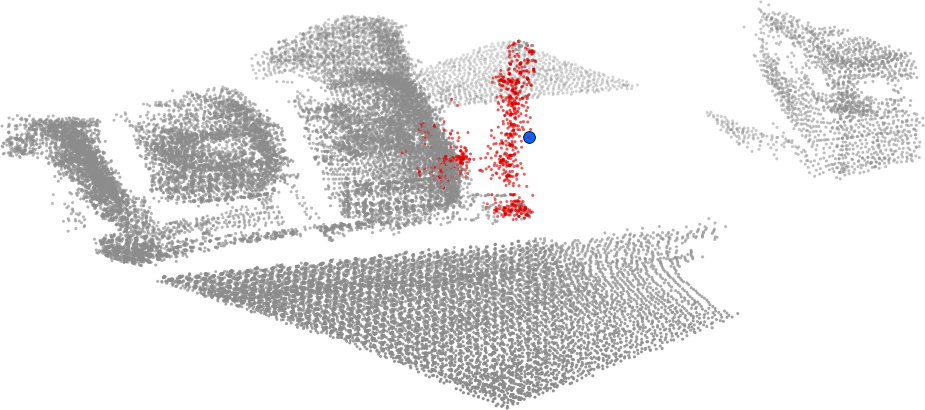}{IoU@1=55.8}
& \QualCellImg{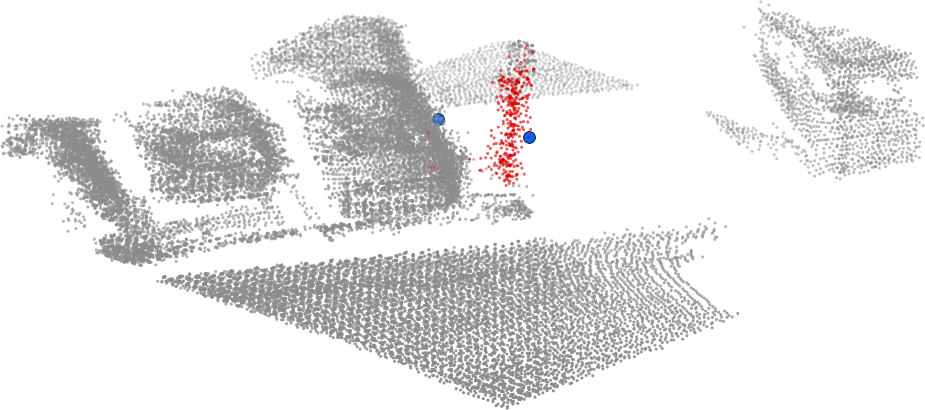}{IoU@2=60.7}
& \QualCellImg{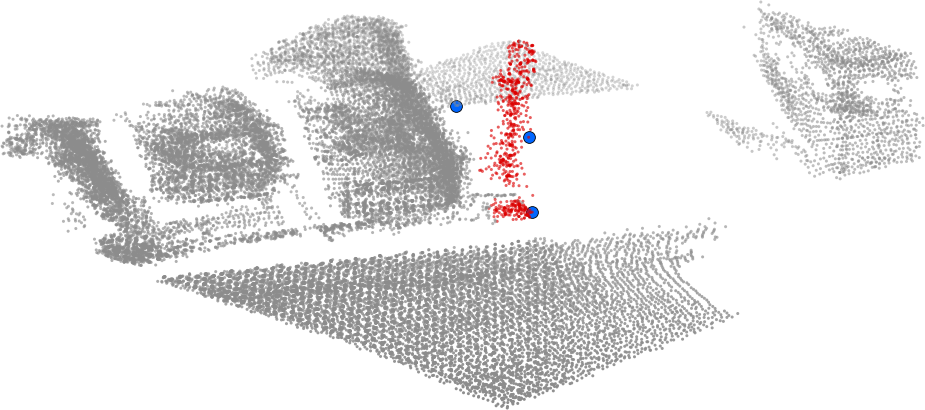}{IoU@3=80.9}\\

\bottomrule
\end{tabular}

\caption{\textbf{Qualitative comparison across datasets.} We compare Easy3D~\cite{simonelli2025easy3d} and \textit{NegROI} under the same simulated-click protocol on ScanNet40, S3DIS, and KITTI-360. Each row shows the target instance (green) and predicted masks (red) after 1/2/3 clicks, with the corresponding IoU@1/2/3 reported below each prediction.}
\label{fig:qual_grid}
\end{figure}

We visualize representative cases across datasets in Fig.~\ref{fig:qual_grid} to illustrate how scene-conditioned negative prompts and click-centric ROI refinement improve interactive segmentation.
Compared to Easy3D~\cite{simonelli2025easy3d}, \textbf{NegROI} better suppresses attached or visually similar distractors and recovers sharper object boundaries under sparse clicks (e.g., 1--3 clicks), consistently yielding higher IoU@k.

% --- Boundary-band FP qualitative grid (no Target column) ---
% Preamble requirements:
% \usepackage{graphicx}
% \usepackage{booktabs}
% \usepackage{array}
% \usepackage{makecell}

\begin{figure}[t]
\centering
\setlength{\tabcolsep}{1pt}
\renewcommand{\arraystretch}{1.0}

\newcommand{\imgw}{0.140\linewidth}
\newcommand{\dscol}{7mm}
\newcommand{\DS}[1]{\rotatebox{90}{\scriptsize\textbf{#1}}}

% (optional) normal cell with text (unused here, but keep if you want)
\newcommand{\QualCellImg}[2]{%
  \makecell[c]{\includegraphics[width=\imgw]{#1}\\[-1.5pt]\scriptsize #2}%
}

% --- no-text cell: image only, keep same height via phantom ---
\newcommand{\QualCellImgNoTxt}[1]{%
  \makecell[c]{\includegraphics[width=\imgw]{#1}\\[-1.5pt]\scriptsize\phantom{Click 3}}%
}

\begin{tabular}{@{}>{\centering\arraybackslash}m{\dscol} c c c c c c@{}}
\toprule
& \multicolumn{3}{c}{\footnotesize\textbf{Easy3D (Band-FP)}} 
& \multicolumn{3}{c}{\footnotesize\textbf{NegROI (Band-FP)}}\\
\cmidrule(lr){2-4}\cmidrule(lr){5-7}
& \scriptsize Click 1 & \scriptsize Click 2 & \scriptsize Click 3
& \scriptsize Click 1 & \scriptsize Click 2 & \scriptsize Click 3 \\
\midrule

% ---------------- ScanNet40 ----------------
\DS{ScanNet40}
& \QualCellImgNoTxt{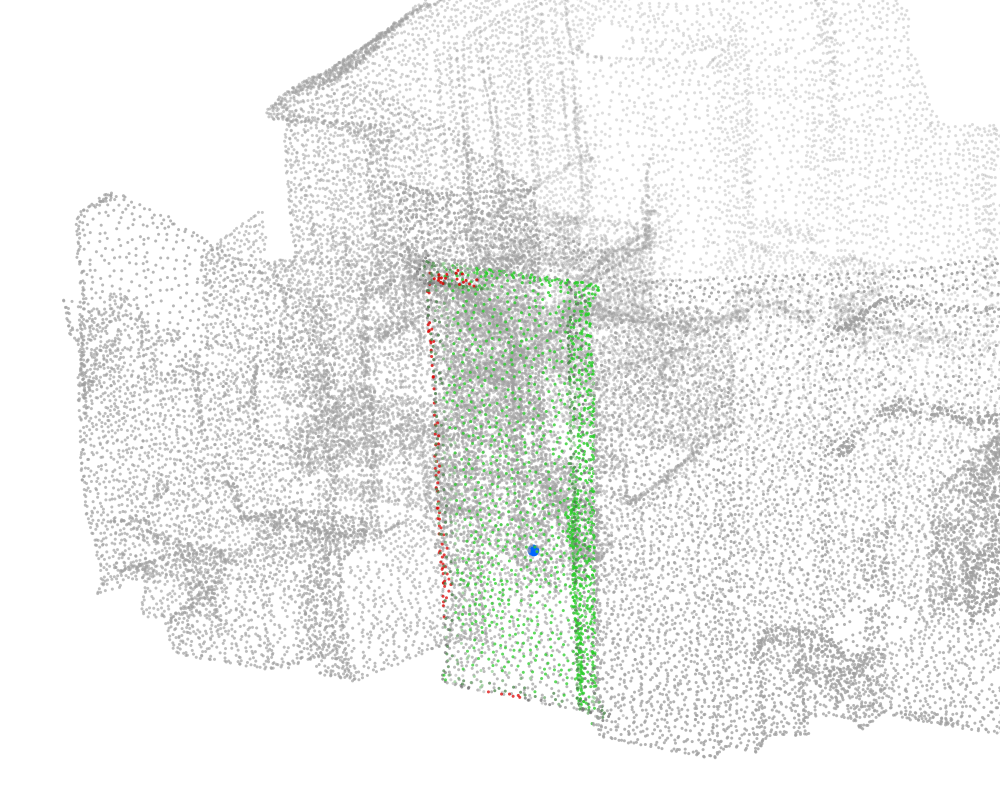}
& \QualCellImgNoTxt{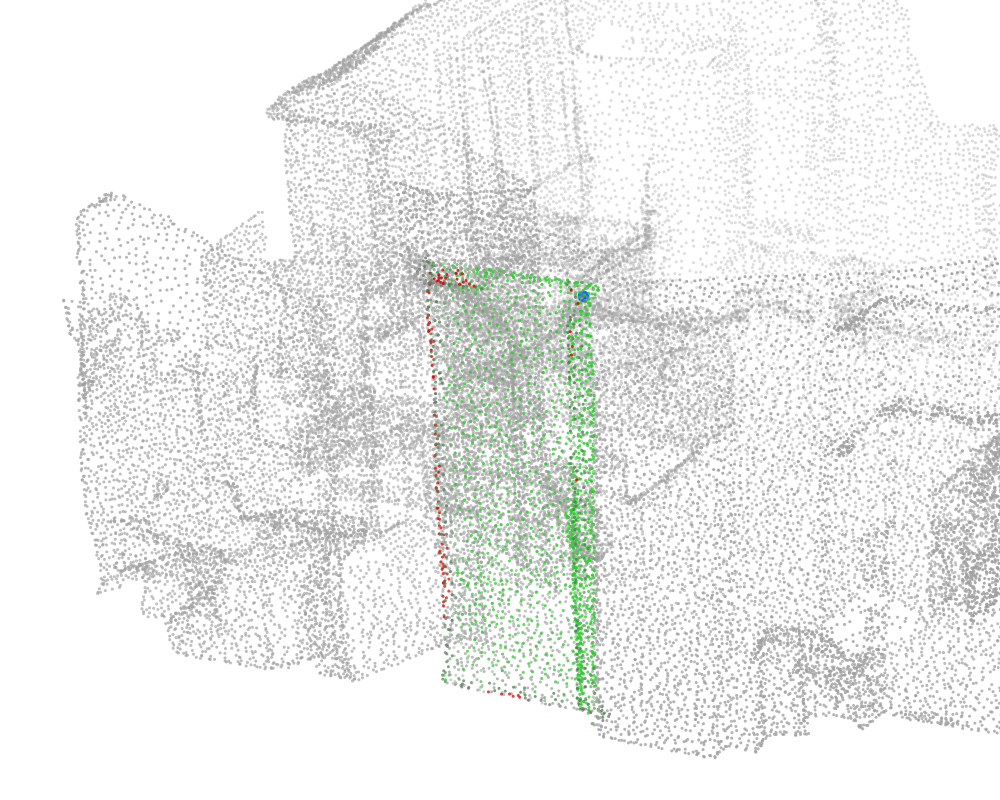}
& \QualCellImgNoTxt{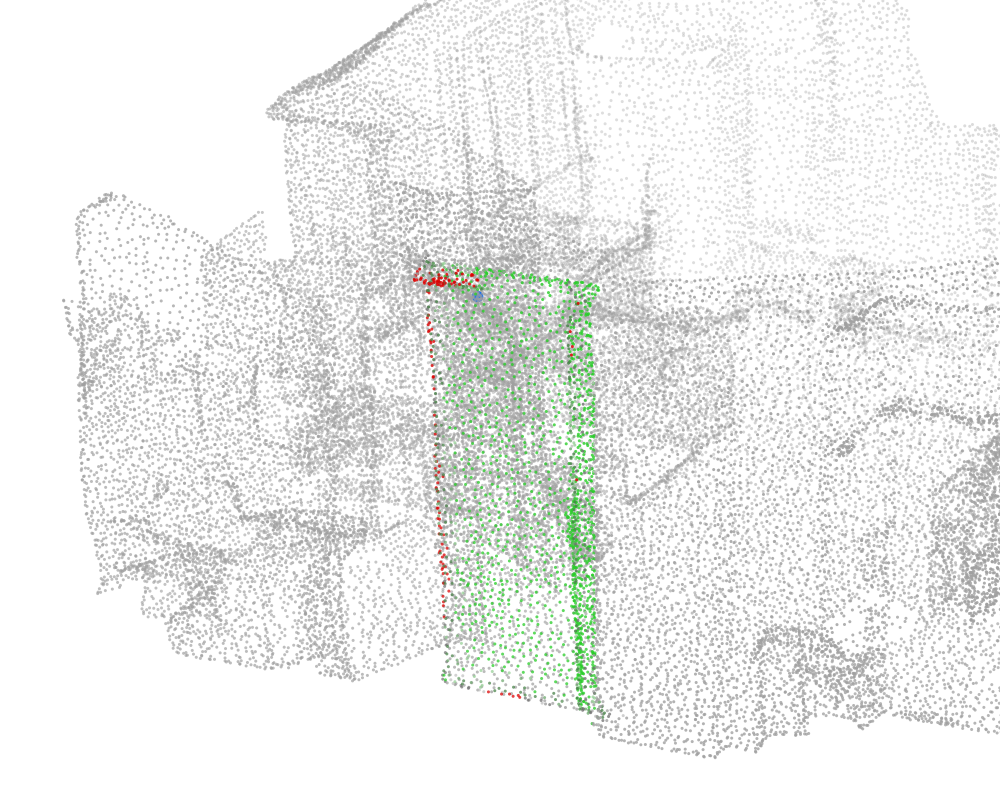}
& \QualCellImgNoTxt{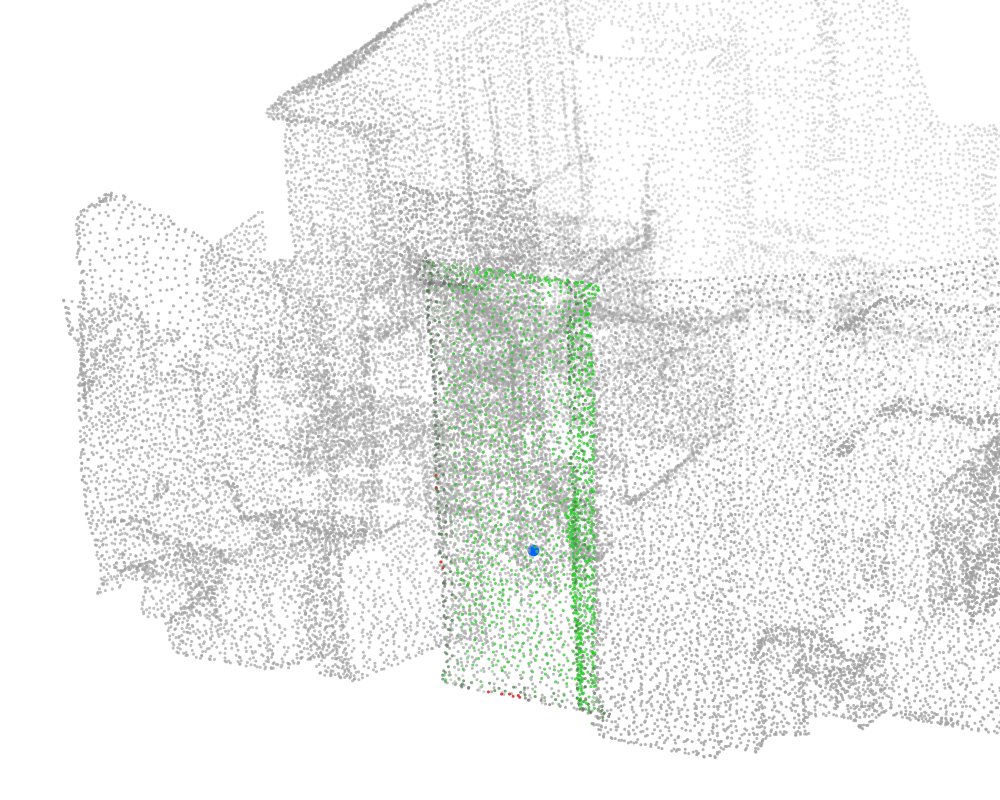}
& \QualCellImgNoTxt{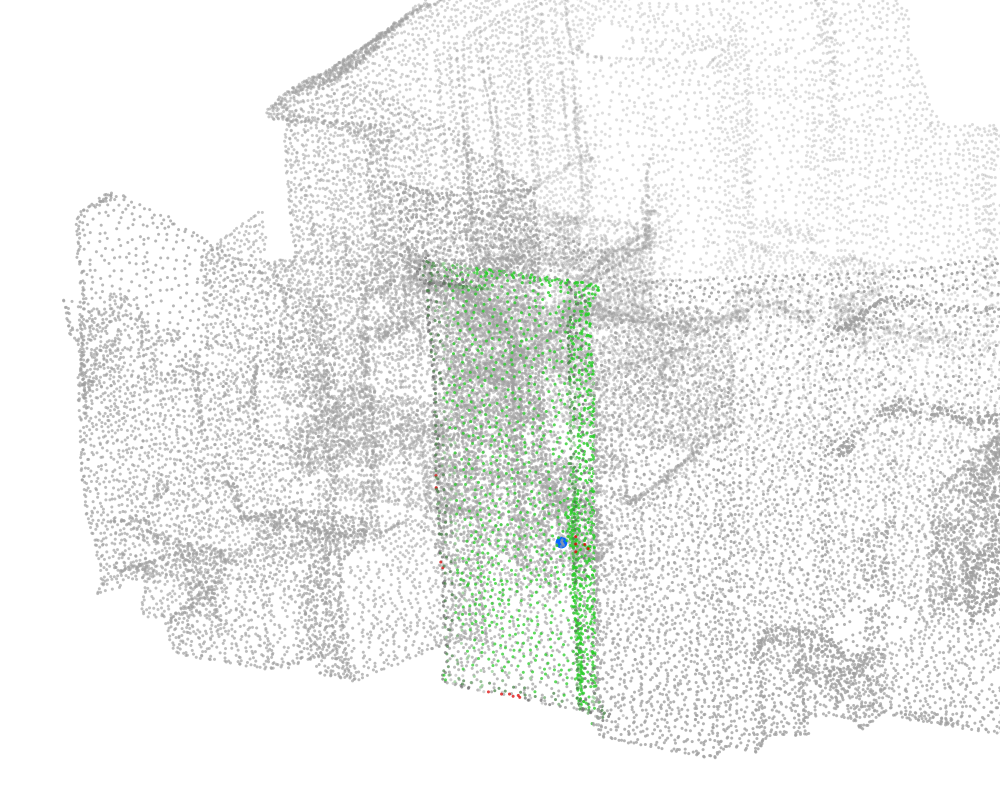}
& \QualCellImgNoTxt{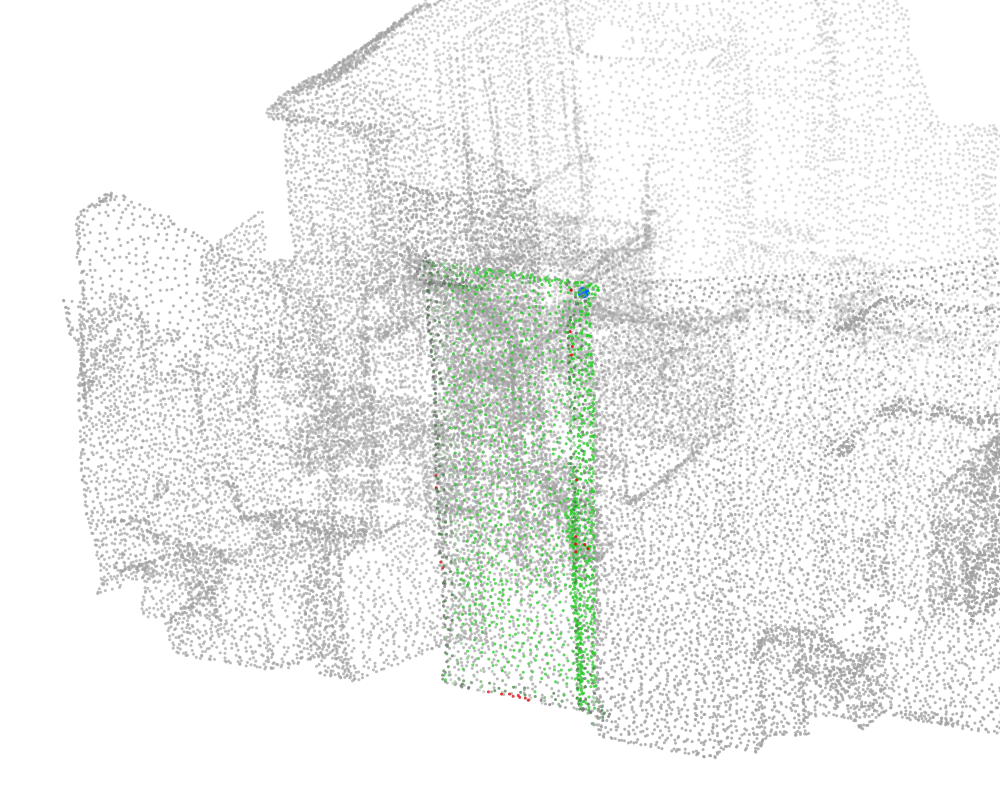}\\
\addlinespace[3pt]

% ---------------- S3DIS ----------------
\DS{S3DIS}
& \QualCellImgNoTxt{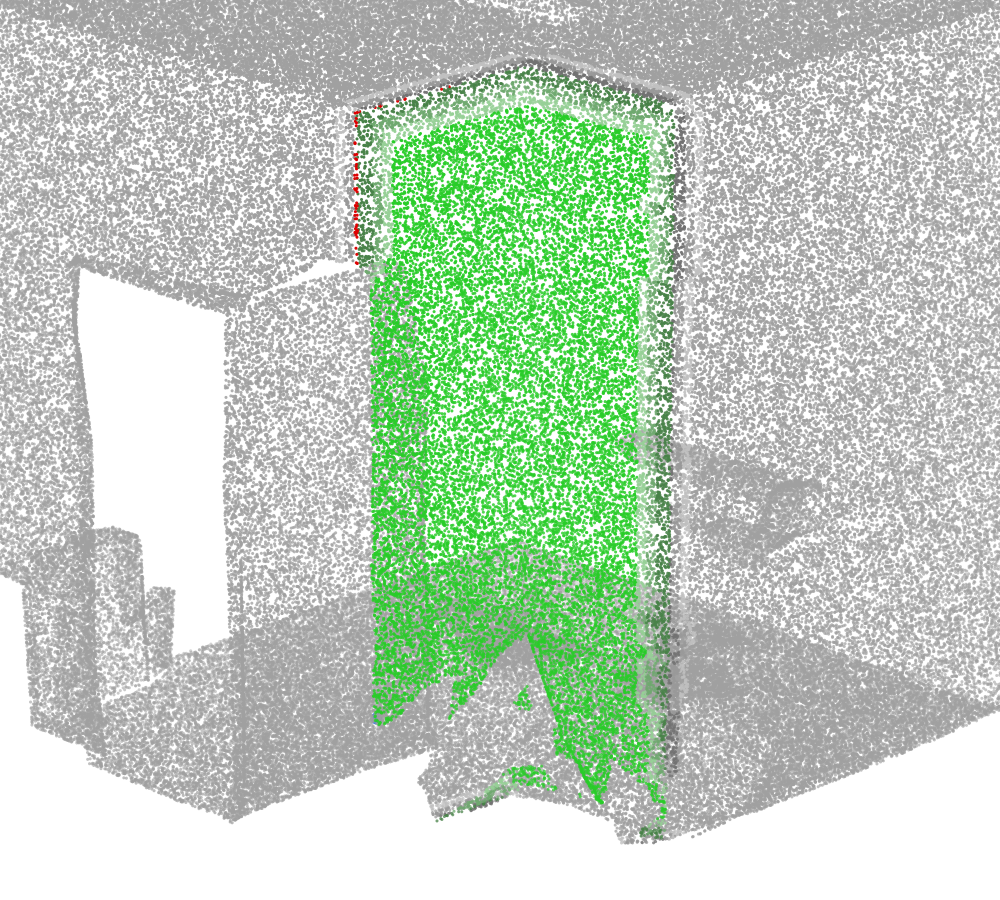}
& \QualCellImgNoTxt{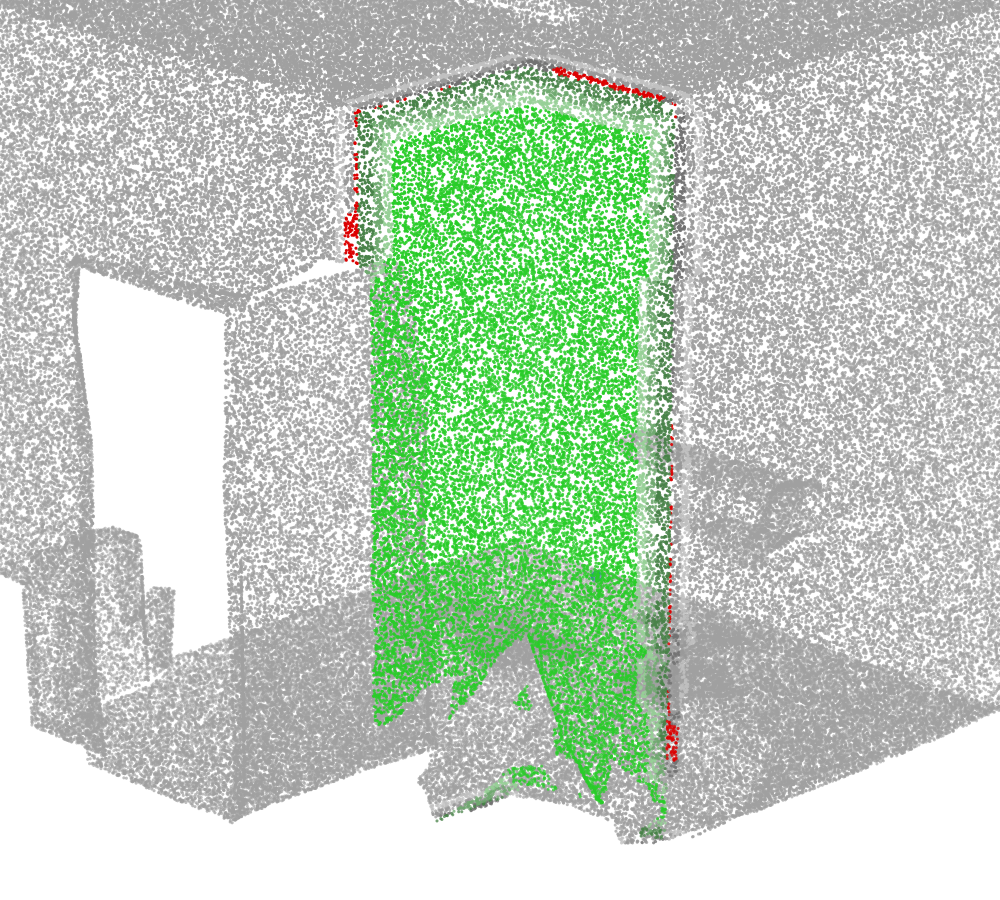}
& \QualCellImgNoTxt{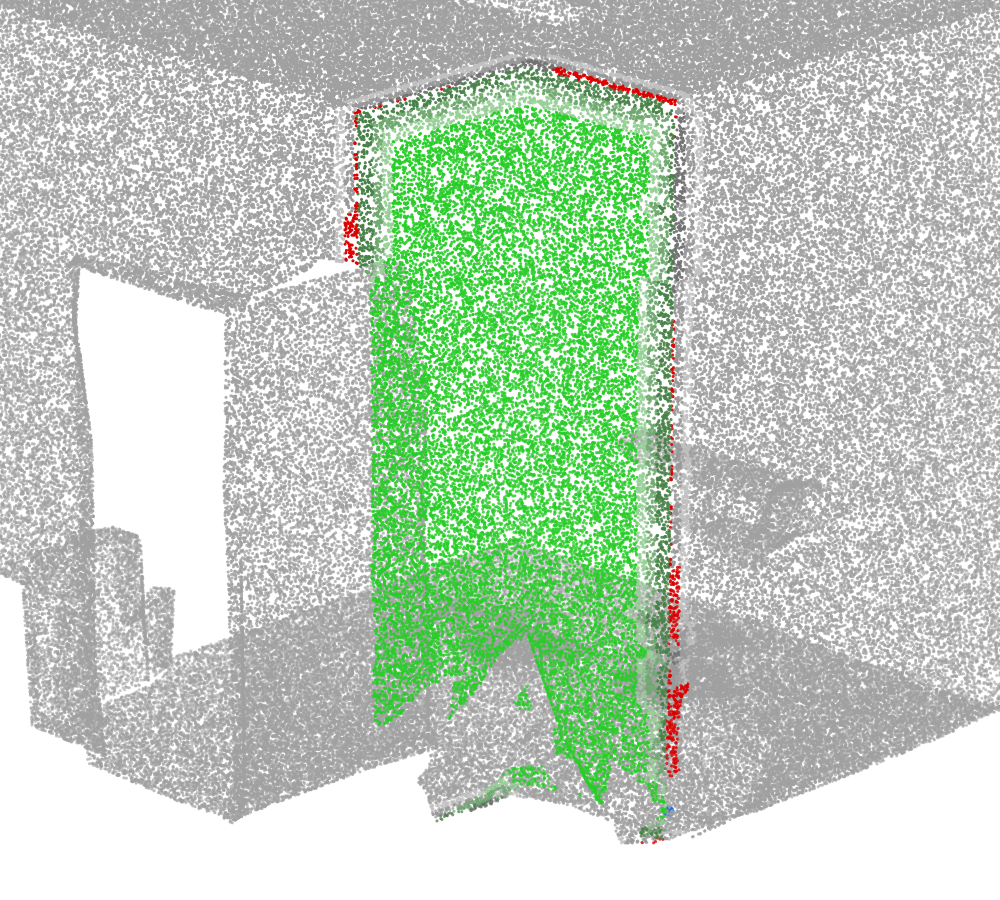}
& \QualCellImgNoTxt{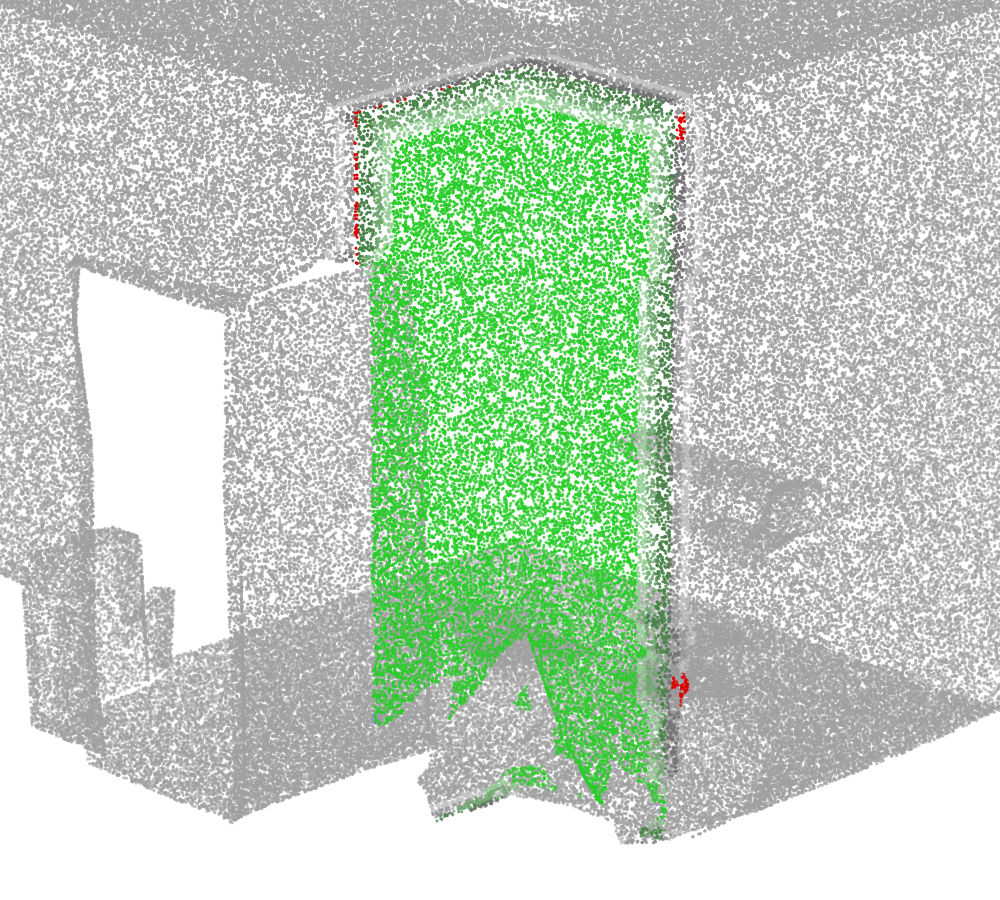}
& \QualCellImgNoTxt{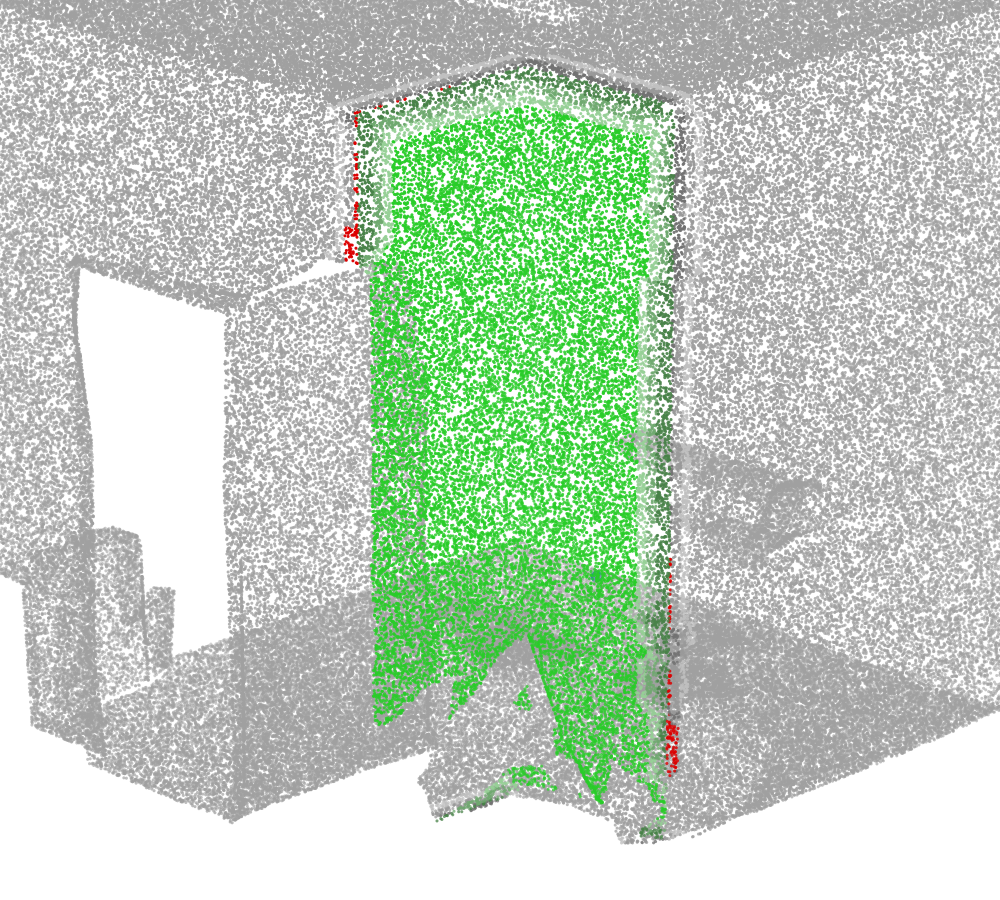}
& \QualCellImgNoTxt{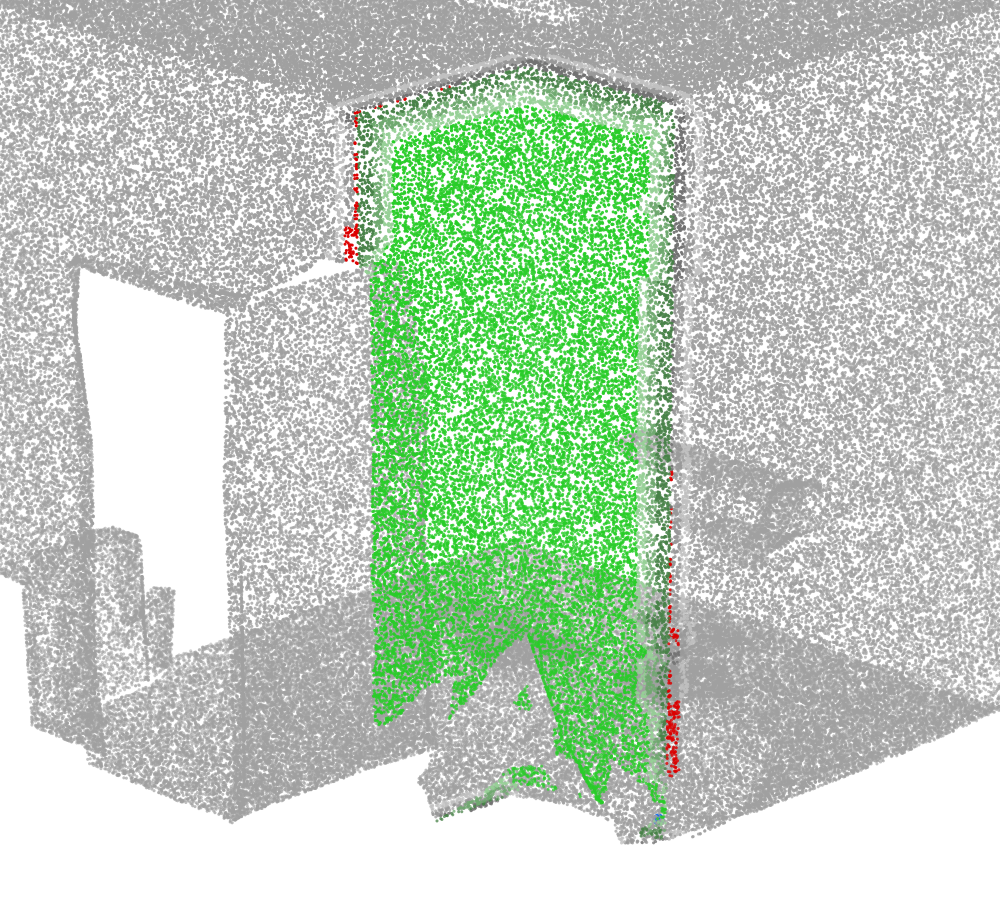}\\
\addlinespace[3pt]

% ---------------- KITTI-360 ----------------
\DS{KITTI-360}
& \QualCellImgNoTxt{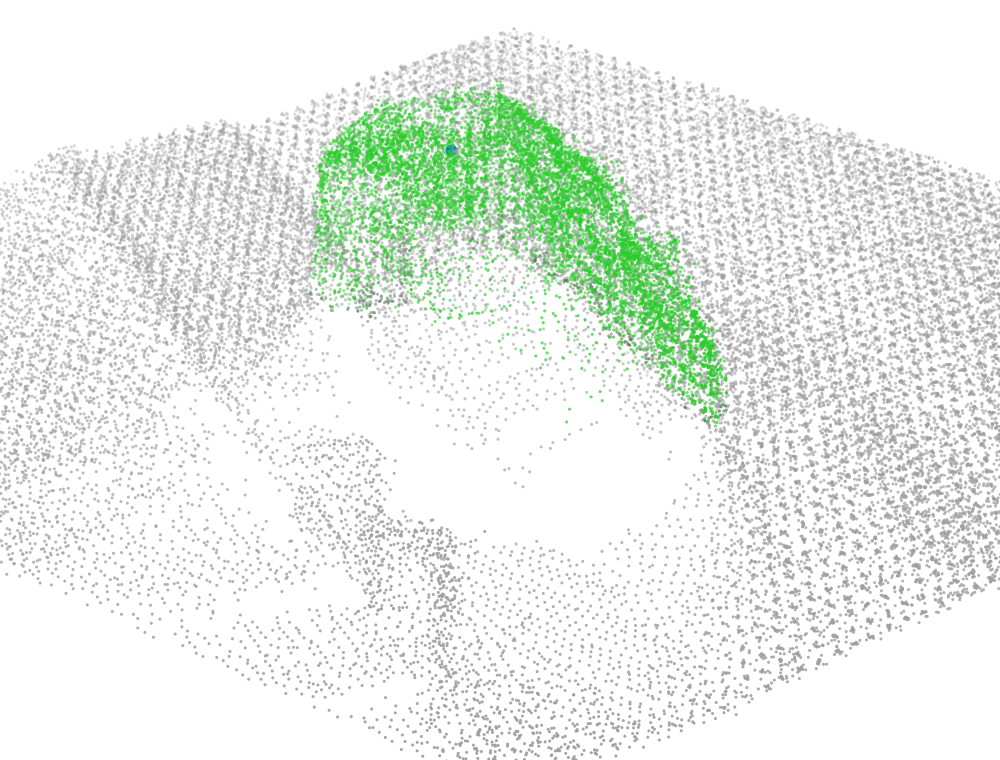}
& \QualCellImgNoTxt{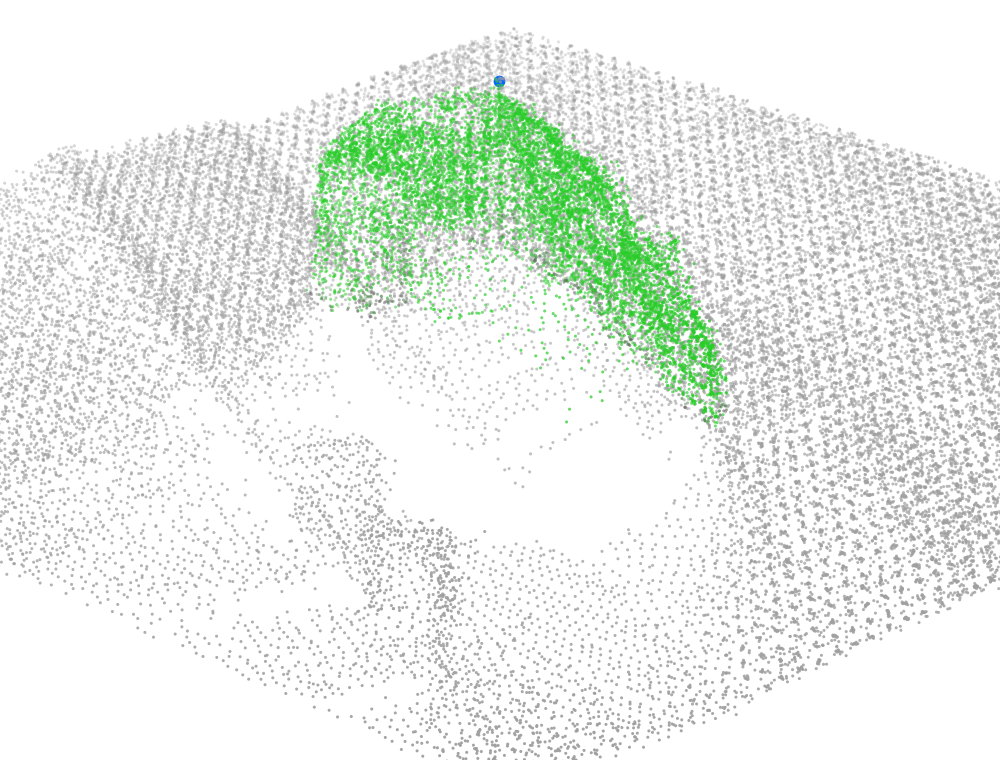}
& \QualCellImgNoTxt{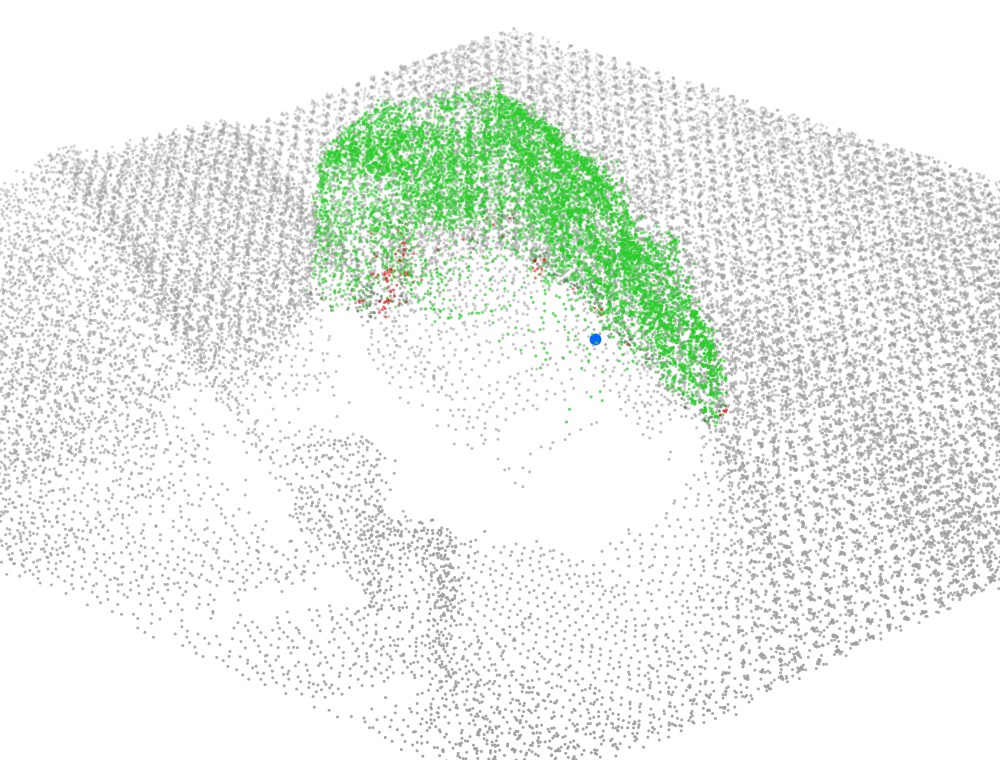}
& \QualCellImgNoTxt{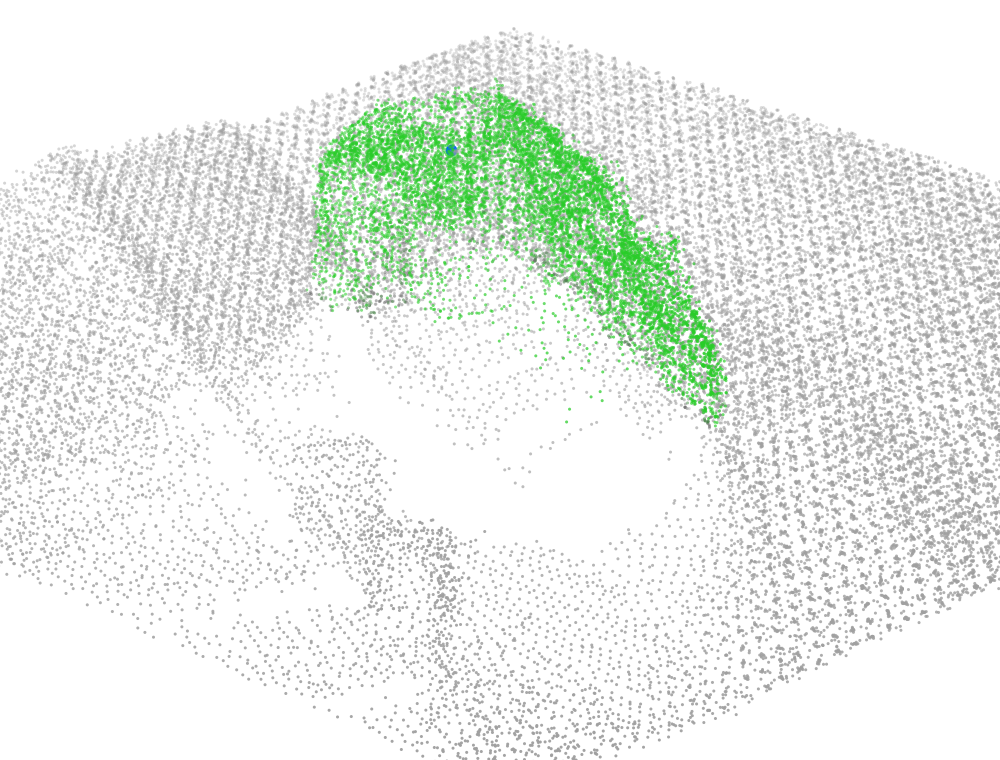}
& \QualCellImgNoTxt{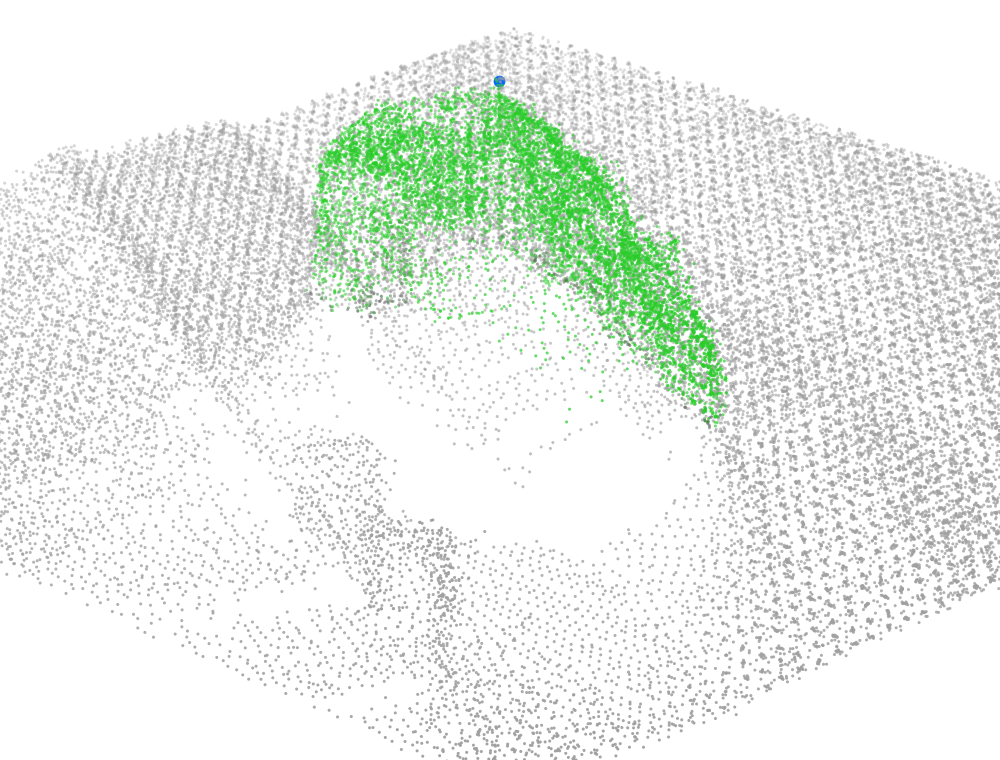}
& \QualCellImgNoTxt{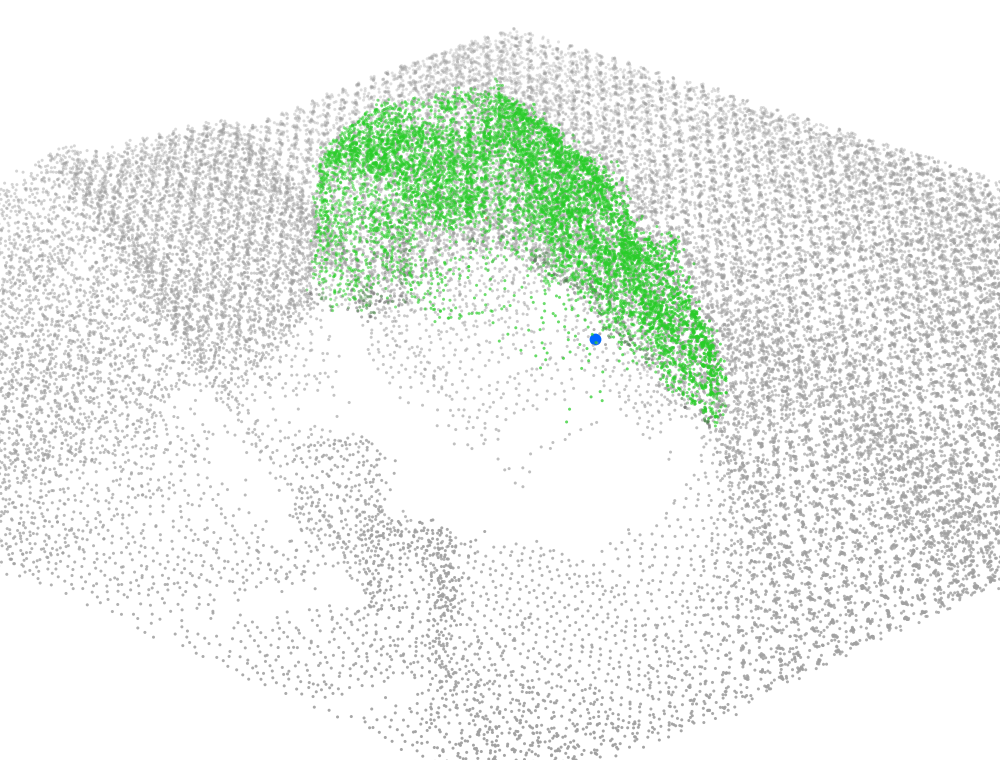}\\

\bottomrule
\end{tabular}

\caption{\textbf{Boundary-band false positive (Band-FP) visualization.}
Gray indicates the ground-truth boundary band (darker near the boundary); red highlights false-positive predictions within the band.
Compared to Easy3D~\cite{simonelli2025easy3d}, \textbf{NegROI} produces fewer boundary-proximal false positives under the same simulated clicks across ScanNet40, S3DIS, and KITTI-360.}
\label{fig:bandfp_grid}
\end{figure}

\paragraph{Boundary-proximal false positives.}
Figure~\ref{fig:bandfp_grid} visualizes \emph{boundary-band false positives} (Band-FP): Gray denotes a narrow ground-truth boundary band (darker is closer to the boundary) and red marks false-positive predictions inside the band.
Across ScanNet40, S3DIS, and KITTI-360, NegROI consistently shows fewer boundary-proximal false positives than Easy3D~\cite{simonelli2025easy3d}, and the red points diminish faster from Click~1 to Click~3.
This trend is consistent with our design: the boundary-aware hard negative loss explicitly guides the negative prompts toward boundary-adjacent confusers (Sec.~\ref{sec:negprompts}), while click-centric ROI refinement sharpens local decision boundaries on a finer grid and fuses them back to the coarse mask (Sec.~\ref{sec:roi_refine}).

% --- Qualitative grid (keep your existing figure code) ---

\section{Conclusions}
We presented a robust interactive 3D segmentation framework that couples click-centric multi-resolution ROI refinement with scene-conditioned negative prompts.
NegROI refines boundaries efficiently near the current click, focuses computation on uncertain regions, and explicitly suppresses hard false positives via boundary-aware negative prompt supervision and diversity regularization.
Experiments on ScanNet, S3DIS, and KITTI-style point clouds indicate improved click efficiency and stronger cross-dataset robustness.
Future work includes fully differentiable ROI selection and stronger test-time adaptation driven solely by user clicks.

\section*{Acknowledgments}
We thank all reviewers and area chairs for their insightful comments and valuable suggestions. We also thank Yueyang Wen for the fruitful discussions and sharing source codes. This work was supported in part by Major Research Plan of the National Natural Science Foundation of China (92048301), Anhui Provincial Major Research and Development Plan (202004H07020008), and Anhui Provincial Natural Science Foundation (2208085MF172).

% ------------------------------------------------------------
% Bibliography (add your related work citations there)
% ------------------------------------------------------------

\FloatBarrier
\bibliographystyle{splncs04}
\bibliography{references}

\end{document}